\documentclass[10pt,twocolumn,letterpaper]{article}

\usepackage{cvpr} 

\usepackage{graphicx}
\usepackage{amsmath}
\usepackage{amssymb}
\usepackage{booktabs}

\usepackage[misc]{ifsym}

\newcommand\blfootnote[1]{
  \begingroup
  \renewcommand\thefootnote{}\footnote{#1}
  \addtocounter{footnote}{-1}
  \endgroup
}

\usepackage[pagebackref,breaklinks,colorlinks]{hyperref}

\usepackage[capitalize]{cleveref}
\crefname{section}{Sec.}{Secs.}
\Crefname{section}{Section}{Sections}
\Crefname{table}{Table}{Tables}
\crefname{table}{Tab.}{Tabs.}

\def\cvprPaperID{2849} 
\def\confName{CVPR}
\def\confYear{2022}

\begin{document}

\title{Structured Sparse R-CNN for Direct Scene Graph Generation}

\author{
  Yao Teng \quad \quad Limin Wang\textsuperscript{~\Letter}\\
  State Key Laboratory for Novel Software Technology, Nanjing University, China\\
  \small \texttt{tengyao19980325@gmail.com}, \texttt{lmwang@nju.edu.cn} \\
}
\maketitle

\begin{abstract}
   Scene graph generation (SGG) is to detect object pairs with their relations in an image.  Existing SGG approaches often use multi-stage pipelines to decompose this task into object detection, relation graph construction, and dense or dense-to-sparse relation prediction. Instead, from a perspective on SGG as a direct set prediction, this paper presents a simple, sparse, and unified framework, termed as Structured Sparse R-CNN. The key to our method is a set of learnable triplet queries and a structured triplet detector which could be jointly optimized from the training set in an end-to-end manner. Specifically, the triplet queries encode the general prior for object pairs with their relations, and provide an initial guess of scene graphs for subsequent refinement. The triplet detector presents a cascaded architecture to progressively refine the detected scene graphs with the customized dynamic heads. In addition, to relieve the training difficulty of our method, we propose a relaxed and enhanced training strategy based on knowledge distillation from a Siamese Sparse R-CNN. We perform experiments on several datasets: Visual Genome and Open Images V4/V6, and the results demonstrate that our method achieves the state-of-the-art performance. In addition, we also perform in-depth ablation studies to provide insights on our structured modeling in triplet detector design and training strategies. The code and models are made available at \url{https://github.com/MCG-NJU/Structured-Sparse-RCNN}.
\end{abstract}
\blfootnote{\Letter: Corresponding author.}

\section{Introduction}

\begin{figure*}
\centering
\begin{subfigure}{0.31\linewidth}
    \centering
    \includegraphics[scale=0.5]{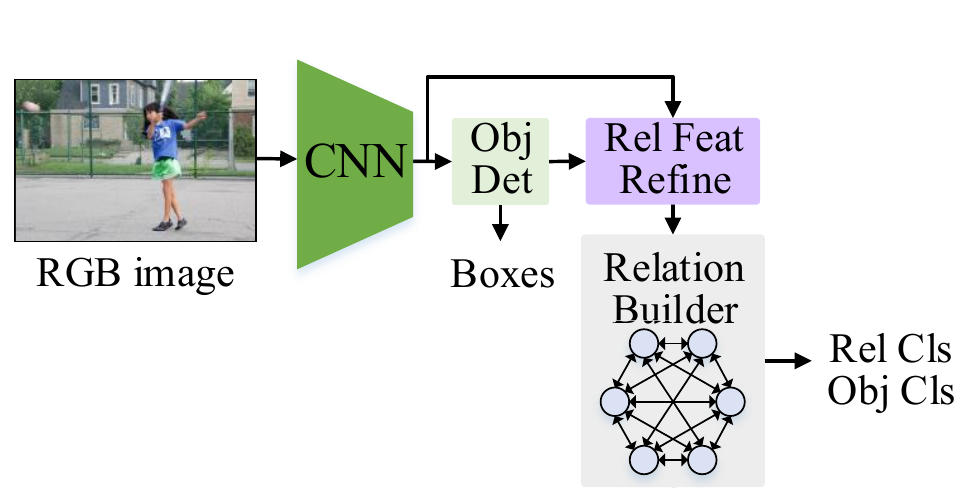}
    \caption{Dense: MOTIFS}
    \label{Fig:example_dense_infer}
\end{subfigure}
\centering
\begin{subfigure}{0.31\linewidth}
    \centering
    \includegraphics[scale=0.5]{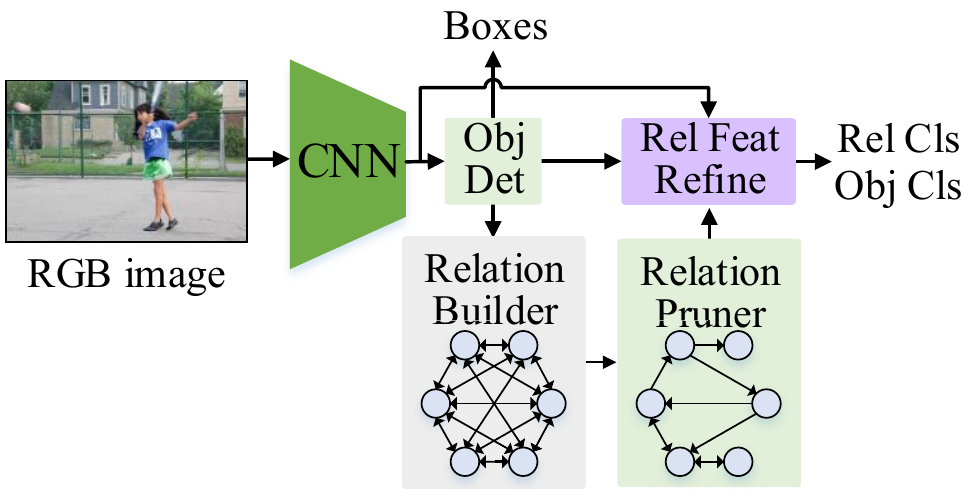}
    \caption{Dense-to-Sparse: Graph R-CNN}
    \label{Fig:example_dense_infer_prune}
\end{subfigure}
\centering
\begin{subfigure}{0.31\linewidth}
    \centering
    \includegraphics[scale=0.5]{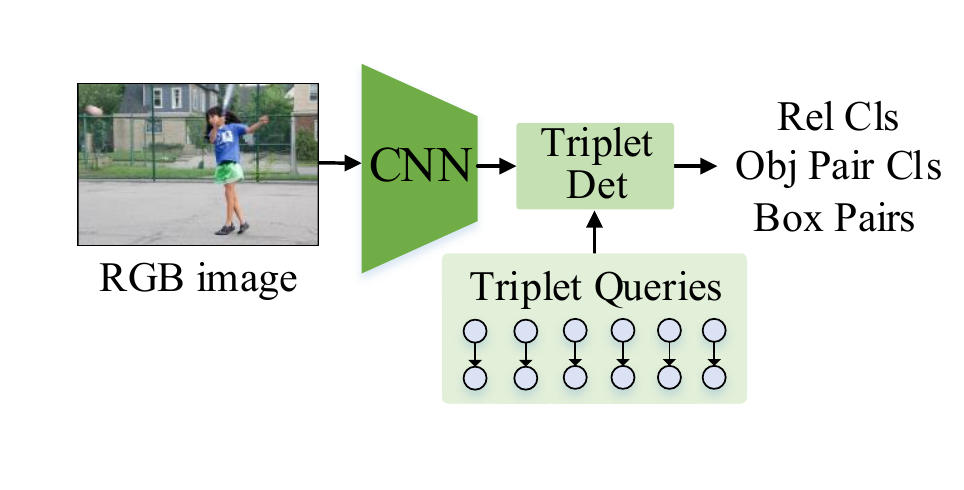}
    \caption{Sparse: Ours}
    \label{Fig:example_query_infer}
\end{subfigure}
\vspace{-2mm}
   \caption{{\bf Comparison of scene graph generation pipeline}. (a) The dense detectors enumerates all object pairs for relation inference, \eg MOTIFS~\cite{neural_motifs}. (b) The dense-to-sparse detectors utilize a pruning scheme to remove unreasonable pairs before the relation inference, \eg Graph~R-CNN~\cite{graphrcnn}. (c) Our network directly generates sparse scene graphs with triplet queries.}
\label{Fig:pipelines}
\vspace{-3mm}
\end{figure*}

Scene graph generation~(SGG)~\cite{Scene_graph_generation_by_iterative_message_passing} aims at detecting objects with their pairwise relations in an image. This structured representation could serve as an effective and compact representation for high-level visual understanding tasks such as image captioning~\cite{caption_cite1,caption2} and visual question answering~\cite{block_vqa,vqa2,vqa3}. Structure information between visual entities is the key to the success of many SGG methods. To capture this structure information, most existing methods typically follows a multi-stage pipeline to decompose this complex task into sub-tasks of object detection, fully-connected relation graph construction, dense relation classification~\cite{neural_motifs,tang_treelstm,reldn}, or dense-to-sparse relation classification~\cite{graphrcnn}, as shown in~\cref{Fig:pipelines}. These well-established methods often rely heavily on object detection performance and involve redundant computation for fully-connected relation graph construction.

\begin{figure}
    \centering
    \includegraphics[width=0.7\linewidth]{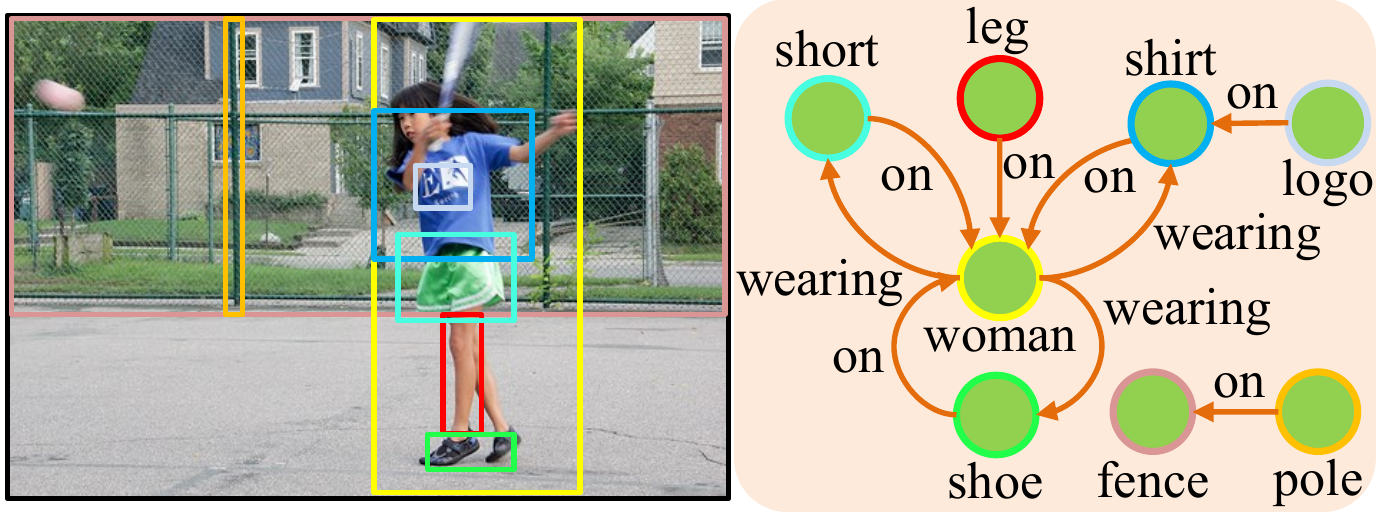}
    \vspace{-3mm}
    \caption{An example of scene graph generation. The scene graph is relatively sparser than the fully connected graph.}
    \label{Fig:example_introduce}
    \vspace{-3mm}
\end{figure}

In addition to structure information, we observe that sparsity is another important property on relation detection in natural images. For example, in~\cref{Fig:example_introduce}, the ground-truth triplets of $\langle$~{\em leg, on, woman}~$\rangle $ and $\langle$~{\em logo, on, shirt}~$\rangle $ are more commonly expressed than the relation between {\em logo} and {\em leg}.
Most existing dense or dense-to-sparse detection methods for SGG fails to well capture the general sparse and semantic priors.
Accordingly, inspired by the recent sparse object detectors (\eg DETR~\cite{detr}, Sparse R-CNN~\cite{sparsercnn}), we present a new perspective on SGG by treating it as a direct sparse set prediction problem. However, unlike sparse object detection, sparse SGG is much more challenging due to its inherent difficulty in object pairing and relation prediction. 

In this paper, we propose a direct sparse scene graph generation framework without explicit object detection and relation graph construction for inference, coined as Structured~Sparse~R-CNN. As shown in~\cref{Fig:example_query_infer}, the key to our Structured~Sparse~R-CNN is a set of learnable {\em triplet queries} and a structured {\em triplet detector}. These learnable triplet queries, composed of two object boxes, two object content vectors and one relation content vector, are responsible for capturing the general prior for sparse detection and encoding the spatial and appearance information of objects and their relation.
Based on the input of CNN feature maps and triplet queries, our triplet detector progressively detects the visual entities and recognizes their relations. The triplet detector contains two cascaded modules for object pair detection and their relation prediction, respectively.
Specifically, we devise structured connections for each triplet query to capture the hierarchical context information. These structured connections first leverage the local interaction in object pairs (Pair Fusion) for better detection and then utilize the object information (E2R Fusion) for better relation prediction.
The parameters of triplet queries are jointly optimized with network weights.

In practice, we find it is challenging to directly train our Structured Sparse R-CNN from scratch. The major challenge comes from the relatively sparse annotations of relations in the current datasets.
The sparse relation annotations contain \textit{few} related object labels, leading to incomplete supervision signal for our object pair detection.
Furthermore, the negative samples are hard to define in the object pair level.
To solve this issue, we propose to build a Siamese Sparse R-CNN to guide the training of our Structured Sparse R-CNN in a knowledge distillation framework~\cite{distill}. This Sparse R-CNN only generates pseudo-labels~\cite{defense_pseudo_label,crest,unbias_teacher_objdet} for training and is inactivated during testing. With the help of these pseudo-labels, we design a new relaxed matching criteria for set prediction loss and enable the training of Structured Sparse R-CNN to be more stable.
Finally, to deal with imbalance distribution of object and relation categories in datasets, we propose an adaptive focusing parameter in our focal loss and utilize a post-hoc logit adjustment, to further boost the performance.

To verify the effectiveness of our framework, we perform experiments on several datasets: Visual Genome~\cite{vg} and Open Images V4/V6~\cite{openimage}.
The experiment results demonstrate that our model is able to yield new state-of-the-art performance under setting of the same backbone on all datasets.
In addition, we conduct in-depth ablation studies to verify the effectiveness of structure modeling in our design. In summary, our main contribution is threefold:

\begin{itemize}
    \item We present a new sparse and unified framework for direct scene graph generation, without explicit object detection and preceding graph construction for inference. This new framework equipped with the structured connection proposed by us shares several advantages, namely simplicity without multi-stage design, effective context modeling, and high efficiency.
    \item We present a practical training strategy to overcome the training difficulty of Structured Sparse R-CNN.
    The knowledge distilled from a Siamese Sparse R-CNN can generate useful pseudo-labels to guide our training. We also propose an adaptive focusing parameter and utilize logit adjustment for imbalance distribution of objects and relations.
    \item Experiment results demonstrate that our simple framework is able to yield the state-of-the-art performance for scene graph generation on Visual Genome and Open Images V4/V6. We also perform detailed ablation studies to provide insights on our designs. 
\end{itemize}

\begin{figure*}
\begin{center}
\includegraphics[width=0.81\linewidth]{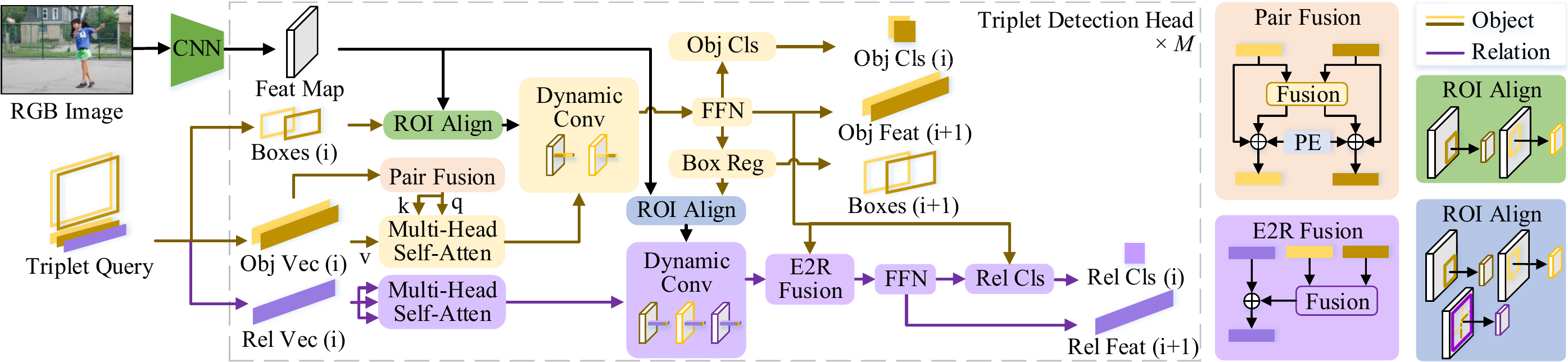}
\end{center}
\vspace{-4mm}
   \caption{
   \textbf{Structured Sparse R-CNN.}
   Our method presents a simple, sparse and unified framework for direct scene graph generation without explicit object detection and relation graph construction in advance. Our framework is composed of CNN backbone, triplet queries, and triplet detector. The triplet queries encode the prior information on object boxes, object appearance, and relation appearance. The triplet detector consists of a series of detection heads. The detector takes CNN features and triplet queries as input, and progressively refine the relation detection results with two cascaded modules (marked in yellow and purple). The vectors of triplet queries are jointly optimized with the network weights with back-propagation.
   (i) in this figure denotes the index of current head. PE denotes positional encoding.
   }
\label{Fig:triplet_det_thread}
\vspace{-2mm}
\end{figure*}

\section{Related Work}
\noindent
\textbf{Scene Graph Generation (SGG).}
In this part, we will discuss the existing works for SGG from three aspects: relation modeling, pipeline and long-tailed distribution.
The explicit modeling for relations~\cite{Scene_graph_generation_by_iterative_message_passing,graphrcnn,wangwenbin2,F-net,Bipartite_Graph_imbalance_rel,vip,Teng0LW21} is commonly considered.
Xu \etal~\cite{Scene_graph_generation_by_iterative_message_passing} built a bipartite graph composed of object proposals as object nodes and union proposals as relation nodes, and the message was passed between them to emphasize features.
Yang \etal~\cite{graphrcnn} utilized the pairwise object features to select few candidates of relation nodes in GCN~\cite{gcn} for classification.
Since MOTIFS~\cite{neural_motifs} was proposed, many works~\cite{tang_treelstm,reldn,gps,Sketching_Image_Gist,unbias_sgg,neural_motifs} begin to aggregate context information for relation classification into object nodes, and the explicit relation features are only treated as attachments.
As for the pipeline, almost all previous works revolve around the concept of multi-stage relation detection and ignore the feasibility of the one-stage paradigm.
Recently, some works started to focus on the one-stage relation detection~\cite{pixel_graphs,FullySGG,qpic_hoi,e2e_hoi,setprediction_hoi,hotr_hoi,mining_onestage_hoi}, but almost all of them still consider explicit post-hoc object detection to boost the performance, thereby making the overall framework a bit complicated. Some of them even do not make full use of the sparse and semantic priors.
As for the long-tailed relation distribution, Tang \etal~\cite{unbias_sgg} performed a variant of logit calibration based on causal analysis. Li \etal~\cite{Bipartite_Graph_imbalance_rel} studied the re-sampling approach for SGG.
In this paper, we directly generate a graph based on a sparse set of queries as its basic elements with efficiency and accuracy. We also propose a corresponding training strategy to get rid of explicit object detection when performing inference. In addition, we revisit the explicit modeling of relations and utilize logit adjustment~\cite{logit_adjustment} for the long-tailed datasets.

\noindent
\textbf{Sparse Object Detector.} Recently, numerous works for sparse object detection were proposed.
DETR~\cite{detr} uses the Hungarian loss~\cite{detr} and a transformer~\cite{transformer} architecture for object detection based on few queries as sparse anchors.
Deformable~DETR~\cite{deformable_detr} boosts the performance by combining the deformable convolution~\cite{deformable} with the transformer and utilizing multi-scale features.
Sparse~R-CNN~\cite{sparsercnn} is more lightweight than these methods and easier to serve as a baseline.
In this paper, we extend Sparse~R-CNN into our sparse triplet detector for generalized relation detection, and design corresponding structures as well as specific training strategy for sparse SGG.

\section{Proposed Approach}

\textbf{Overview.} Unlike the previous SGG methods composed of multiple stages, our Structured~Sparse~R-CNN presents a simple, direct and unified framework for relation detection. Our method takes image features and a set of triplet queries as inputs, and passes them into the stacked detection heads to progressively detect objects and predict their relations. The parameters of triplet queries can be jointly optimized with network weights in an end-to-end manner. We detail these components in the sequel.

\subsection{Structured Sparse R-CNN}

\textbf{Backbone.}
The image is fed into a convolution neural network~(CNN)~\cite{resnext} with Feature Pyramid Network (FPN)~\cite{fpn} for feature extraction, and then the feature maps are fed into our triplet detector to detect objects and predict relations.
More details can be found in Section~\ref{sec:implement}.

\textbf{Triplet query.}
To localize objects and recognize their categories and relations, our Structured Sparse R-CNN uses a set of learnable triplet queries to represent the general distribution prior of triplets. Specifically, each triplet query is composed of two proposal boxes representing the locations of objects, two object content vectors encoding the appearance of objects, one relation content vector capturing the structure information between objects. Each box is a 4-d parameter to represent the normalized box center, width, and height. The object and relation features are represented by 1024-d and 256-d parameters respectively, which encode the semantics of objects and relations. 

These triplet queries are randomly initialized during training and jointly optimized with network weights via back-propagation algorithm. Once the training is finished, these learnt triplet queries serve as the general prior for SGG and are the same for all testing images. Basically, the learnt triplet queries could be viewed as the general statistics of potential objects location, appearance, and their relations, discovered in a data-driven manner from training set. They provide an initial guess for the triplet candidates, which is then refined progressively with the triplet detector.

\textbf{Triplet detection head.}
Our Structured Sparse R-CNN is composed of a series of modular network building blocks, termed as {Triplet Detection Head}, 
to progressively refine the location and categories of objects as well as the prediction of relations.
As shown in~\cref{Fig:triplet_det_thread}, each head of our triplet detector presents two modules to perform object pair detection and their relation prediction, respectively.
These two modules are cascaded together with structured connections to accomplish the task of SGG.

{\em Object pair detection.} 
Given $N$ triplet queries, triplet detector first uses object feature vectors to perform the global and local information interaction.
The traditional multi-head self-attention mechanism is employed for aggregating global context information into objects.
To better describe the context features within object pairs, we propose a pair fusion module (PF) to relate the object feature vectors by using a multi-layer perception (MLP).
The meaning of this structured connection lies in emphasizing each object feature via utilizing the unique properties of the internal interaction, \eg in one triplet, its subject will be aware of which object to pair with.
Moreover, the relations are unlikely to occur between the same objects.
Therefore, this operation is designed to separate the objects and enhance their semantics.
Its specific process is as follows:
\begin{equation}
\begin{gathered}
X_{p}=\mathrm{ReLU}(\mathrm{LN}({W^s_0} X_s + W^o_0 X_o )),\\ X'_s = X_s + {W^s_1} X_{p} + \mathrm{P}_s,~ X'_o = X_o + {W^o_1} X_{p} + \mathrm{P}_o,
\end{gathered}
\label{Equ:pair_fusion}
\end{equation}
where $X_s$ and $X_o$ denote the subject and object content vectors, respectively.
$ \mathrm{P}_s $ and $ \mathrm{P}_o $ are positional encoding for subjects and objects.
$W^s_0, W^o_0, W^s_1, W^o_1$ and $W_0$ are learnable matrices.
$\mathrm{LN}(\cdot)$ and $\mathrm{ReLU}(\cdot)$ represent the layer normalization~\cite{layernorm} and ReLU activation~\cite{relu}. $X'_s$ and $X'_o$ are used for generating the key and query vectors in self-attention.

Then the enhanced object feature vector is used to attend the RoI pooled feature of each object independently with a {\em dynamic} convolution~\cite{dynamic_filter}, where the kernels for convolution are produced by object feature vectors. Subsequently, a feed-forward network (FFN)~\cite{transformer} with two 
MLPs (\ie, \ {\em cls} and {\em reg} heads) 
is constructed for object box regression and category classification, respectively.

{\em Relation recognition.} After the object pair detection, our triplet detector perform visual relation prediction for each detected object pair.
After performing a similar dynamic convolution on relation-level features from ROI Align~\cite{roialign} with relation vectors, we introduce a bottom-up connection to combine the object-level features with our relation feature vectors.
This bottom-up structured connection is called as visual entities to relation fusion, denoted by E2R Fusion (E2R).
These object-level features are expected to enhance the relation vectors by providing low-level object information via other MLPs:
\begin{equation}
\begin{gathered}
H_{r}=W_x \mathrm{ReLU}(\mathrm{LN}({W^s_{r}} F_s )) + W_y \mathrm{ReLU}(\mathrm{LN}({W^o_{r}} F_o )), \\
F'_{r} = \mathrm{LN}(F_{r} + H_{r} + {W^p_{r}}\mathrm{ReLU}({W^s_{p}} \mathrm{P}_s + {W^o_{p}} \mathrm{P}_o) ),
\end{gathered}
\label{Equ:rel_fusion}
\end{equation}
where $F_s$, $F_o$ and $F_r$ denote the features of the subjects, objects and relations, respectively.
$W^s_{r}$, $W^o_{r}$, $W_{x}$, $ W_{y}$, ${W^s_{p}}$, ${W^o_{p}}$ and ${W^p_{r}}$ are linear transformation matrices.
Finally, a FFN with relation classification head is used to conduct relation prediction with the enhanced relation vectors.

In addition, due to the object feature is helpful for relation prediction~\cite{reldn}, we use object-level features to directly predict the relation categories as another branch.
The final classification comes from the sum of the outputs of the master branch and this branch.

{\bf Discussion.} 
Our Structured Sparse R-CNN is an extension of the original Sparse R-CNN to the structure prediction task. To mitigate the difficulty of relation detection over object detection, our Structured Sparse R-CNN introduces customized structure modeling in our triplet detector.
The key difference with the original Sparse R-CNN is the consideration of structure information.
First, we introduce the pairwise object context information for better object detection.
Second, we model the hierarchical context information between two objects and their relation.
As shown in experiments, this structure modeling is of great importance in our Structured Sparse R-CNN design to accomplish SGG.

\subsection{Learning with Siamese Sparse R-CNN}

Unlike Sparse R-CNN~\cite{sparsercnn}, we observe that it is challenging to directly train our Structured Sparse R-CNN only with the ground-truth triplets.
These triplet annotations cover {\em too few object samples}. However, for our triplet detector, training its object pair detection component requires a large number of object samples.
Therefore, we consider generating some virtual {\em object pairs} as pseudo-labels for training so as to increase the recall of object pair detection.
For this objective, we present a relaxed and enhanced training strategy based on knowledge distillation~\cite{distill} from an extra Sparse R-CNN which can yield a set of such pseudo-labels.
To allocate these pseudo-labels to predicted object pairs in training, we design a two-stage triplet label assignment with specific classification and regression loss.
Even if these pseudo-labels are not in annotations, they could be used to train the object pair detection component under the new label assignment and loss.

\begin{figure}
\begin{center}
\includegraphics[width=1.0\linewidth]{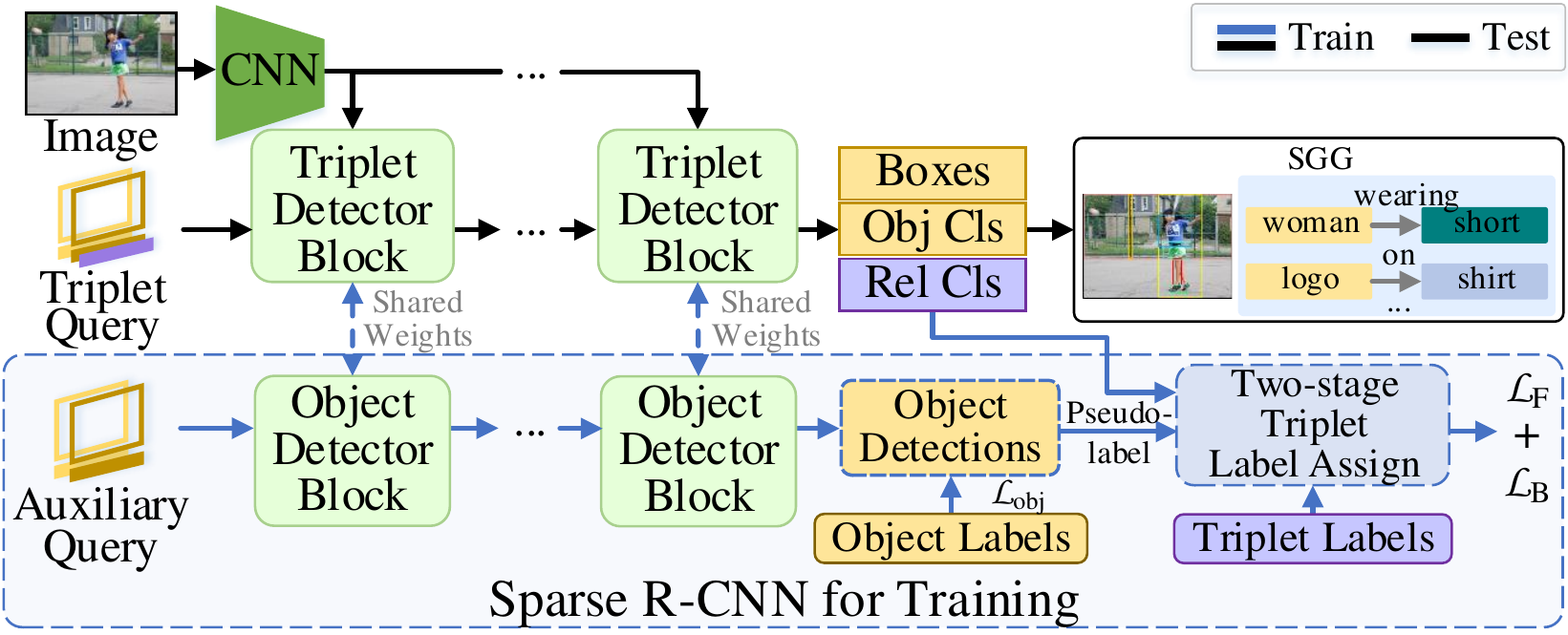}
\end{center}
\vspace{-6mm}
   \caption{
   {\bf Learning with Siamese Sparse R-CNN}. We present a relaxed and enhanced training strategy for our Structured Sparse R-CNN based on the knowledge distillation from a Siamese Sparse R-CNN which is composed of object detection heads. This extra Sparse R-CNN generate pseudo-labels for our triplet label assignment and also augment our triplet queries at each layer, benefiting the training of Structured Sparse R-CNN.
   }
\label{Fig:net_train}
\vspace{-2mm}
\end{figure}

\textbf{Siamese Sparse R-CNN.}
As shown in~\cref{Fig:net_train}, we propose to build an extra Sparse R-CNN {\em only activated in the training phase}. This network shares the same weight with our Structured Sparse R-CNN and thus is called as Siamese Sparse R-CNN.
It is separately employed for object detection, and has the \textit{auxiliary queries} independent of the triplet ones.
It is jointly trained with our triplet detector, and has its own \textit{object label assignment} just like the object detectors~\cite{sparsercnn,detr}. 
The detected objects are grouped into pairs and act as pseudo-labels for training Structured Sparse R-CNN.

\textbf{Two-stage triplet label assignment.}
In training, we first directly use Hungarian matching~\cite{detr} to assign ground-truth relations with their objects to a set of triplet candidates.
Then, for the remaining triplet candidates not matching the ground-truth triplets, instead of padding their objects with background label, we use another Hungarian matching to assign these \textit{object pairs} to a subset of pseudo-labels provided by Siamese Sparse R-CNN.
With such a matching, these triplets are forced to approximate object pairs that most resemble them.
Finally, with the two-stage label assignment, we compute the loss for triplet detection as the sum of $\mathcal{L}_{F}$ for the triplets matched in the first stage and $\mathcal{L}_{B}$ for the ones matched in the second stage. 

In the first stage, a bipartite matching is conducted between ground-truth triplets and all predicted triplets~\cite{hungarian}. 
The following is the matching cost between a prediction and a ground-truth triplet, as well as a part of the final loss:
\begin{equation}
\begin{gathered}
\mathcal{L}_{F} = \lambda_{{cls}_r} \mathcal{L}^g_{{cls}_r}  +  \sum_{i \in \{s, o\}} \lambda_{{cls}_i} \mathcal{L}^g_{{cls}_i} \\ ~~~~ + \lambda_{{L_1}_i} \mathcal{L}^g_{{L_1}_i} + \lambda_{{giou}_i} \mathcal{L}^g_{{giou}_i},
\end{gathered}
\label{Equ:L_FG_t}
\end{equation}
where $\mathcal{L}^g_{{cls}_{i}}$ and $\mathcal{L}^g_{{cls}_r}$ are focal loss~\cite{focalloss} between ground-truth and predicted labels of objects and relations, respectively.
$s$/$o$ refers to the subject/object in one object pair.
$\mathcal{L}^g_{{L_1}_{i}}$ and $\mathcal{L}^g_{{giou}_{i}}$ are L1 loss and generalized IoU loss~\cite{giou} between the bounding boxes of objects and the corresponding ground-truth boxes, respectively.
$\lambda_{{cls}_r}$, $\lambda_{{cls}_i}$, $\lambda_{{L_1}_i}$ and $\lambda_{{giou}_i}$ are the coefficients of each component.

In the second stage, for the set of pseudo-labels, we remove some of its pairs that detect the ground-truth, and denote the remaining pairs as the pseudo-label set $U$.
In principle, we could directly use $U$ to train our triplet detector, but some works show that the hard-label format benefits training~\cite{distill_vit}.
Therefore, considering the objects in $U$ are also assigned with labels during the previous object label assignment, we keep the boxes of the objects {\em not} matching ground-truth objects {\em \textbf{unchanged}} and replace all the classification scores as well as other predicted boxes with the assigned labels.
Then, we perform another bipartite matching between $U$ and the object pairs from remaining triplet predictions. 
Due to the existence of predicted boxes, this stage of label assignment is like distillation.
The matching cost between a predicted pair and a pair in $U$ is as follows:
\begin{equation}
\begin{gathered}
\mathcal{L}^m_{B} = \sum_{i \in \{s, o\}} \eta_{{L_1}_i} \mathcal{L}^u_{{L_1}_i} + \eta_{{giou}_i} \mathcal{L}^u_{{giou}_i} + \mathbf{1}^{u}_i \eta_{{cls}_i} \mathcal{L}^{u}_{{cls}_i},
\end{gathered}
\label{Equ:C_BG_t}
\end{equation}
where $\mathcal{L}^{u}_{{cls}_i}$, $\mathcal{L}^{u}_{{giou}_i}$ and $\mathcal{L}^{u}_{{L_1}_i}$ is loss between predicted objects in triplets and the objects in $U$, just like those in~\cref{Equ:L_FG_t}.
$\eta_{{cls}_i}$, $\eta_{{L_1}_i}$ and $\eta_{{giou}_i} $ are coefficients. $\mathbf{1}^{u}_i$ is 1 if the object from $U$ hits the ground-truth, otherwise is 0.

After the bipartite matching of the second stage, we then pad the relation predictions in the remaining triplets with background label.
The loss for these triplets is as follows:
\begin{equation}
\begin{gathered}
\mathcal{L}_{B} = \lambda_{{cls}_r} \mathcal{L}^{-}_{{cls}_r} + \sum_{i \in \{s, o\}} \eta_{{cls}_i} \mathcal{L}^{u}_{{cls}_i} \\ ~~~~~~~~~~~~~~~~~~~~~~~~ + \mathbf{1}^{u}_i  (\eta_{{L1}_i} \mathcal{L}^{u}_{{L_1}_i} + \eta_{{giou}_i} \mathcal{L}^{u}_{{giou}_i}),
\end{gathered}
\label{Equ:L_BG_t}
\end{equation}
where $\mathcal{L}^{-}_{{cls}_r}$ is focal loss between relation prediction and background label, the other terms are same with the above.

\begin{figure}
    \centering
    \includegraphics[width=0.8\linewidth]{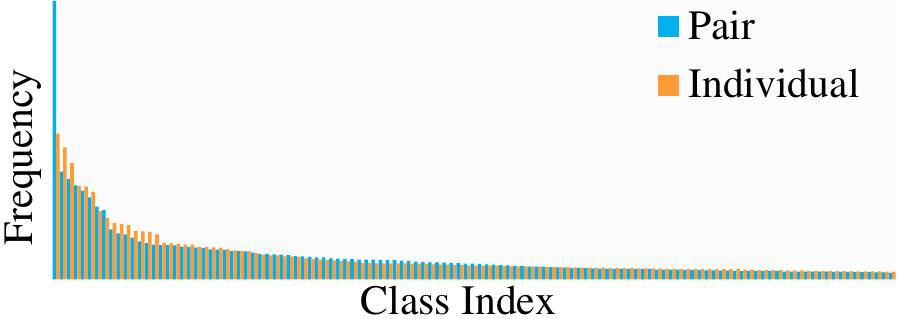}
    \vspace{-2mm}
    \caption{The class distributions of objects as individuals and pairs. The frequency from the two distributions is sorted in descending order separately.}
    \label{Fig:freq_count}
    \vspace{-3mm}
\end{figure}

\subsection{Imbalance Class Distribution}
\textbf{Adaptive focusing parameter.}
The format of triplets may deteriorate the imbalanced class distribution of entities.
As shown in~\cref{Fig:freq_count}, the most frequented class of entities as the elements of pairs in distribution is far heavier than as individuals, which is attributed to the duplicates of entities induced by the format of triplets.
Thus, we consider reducing the weights for majority classes in the loss for object classification.
Inspired by~\cite{gps}, we re-balance the biased model by tailoring the focusing parameter (denoted as $\gamma$) in focal loss~\cite{focalloss} for each category:
\begin{equation}
\begin{gathered}
\gamma(c) = \min \{ 2, 3 - (1 - f_c)^\mu  (-\log(f_c))^{\frac{1}{\mu}} \},
\end{gathered}
\label{Equ:gamma}
\end{equation}
where $c$ denotes the object category. $f_c$ denotes the frequency of each object category occurring in triplets. $\mu $ is a hyper-parameter.

\textbf{Logit adjustment (LA).}
As for the imbalance class distribution of relations, we utilize logit adjustment~\cite{logit_adjustment}. We directly calculates the frequency for each relation category, and the final classification score is obtained from the logit minus the frequency multiplied by a tuning parameter $\tau$.

\begin{table*}
\centering
\small
\setlength{\tabcolsep}{3pt}
\begin{tabular}{ccc|cc|cc|cc|cc}
\toprule
Rel & E2R & PF  & R@20 & R@100 & zR@20  & zR@100 & mR@20  & mR@100  & zR@100 (LA)   & mR@100 (LA) \\
\midrule
 &   &  & 23.16   & 34.62  & 0.79 & 2.44  & 5.09 &  8.62  &  3.03  &  19.04 \\
 & & \checkmark &   25.33  & 36.57  & 0.94 & 2.41 & 5.67 &  9.13   & 2.96 & 20.25 \\
\checkmark &  & \checkmark & 25.52  &  {36.58}  &  {1.38}  &  {3.41} &  6.03  &  9.83   &  3.61 &  19.96    \\ 
 \checkmark & \checkmark &    & 24.32   & 35.34  & 1.24 & 3.32  & 5.77 &  9.37  & 3.83  & 19.55 \\
\checkmark & \checkmark & \checkmark & \textbf{25.82}   & \textbf{36.93}  & \textbf{1.51}  & \textbf{3.74} & \textbf{6.08}  & \textbf{10.04}   & \textbf{4.04} & \textbf{21.39} \\
\bottomrule
\end{tabular}
\vspace{-2mm}
\caption{Study on the structured modules. Rel: using relation feature vectors, PF: pairwise fusion,  E2R: visual entities to relation fusion.}
\vspace{-1mm}
\label{Tab:interaction}
\end{table*}

\begin{table*}
\centering
\small
\setlength{\tabcolsep}{3pt}
\begin{tabular}{cc|cc|cc|cc|cc|c}
\toprule
TLA & NMS & R@20 & R@100 & zR@20  & zR@100 & mR@20  & mR@100   & zR@100 (LA)   & mR@100 (LA) & speed \\
\midrule
full BG &    & 24.62   & 35.00   & 1.21 & 3.11  & 5.75   & 9.20 &  3.52   &  18.57  & 0.19 \\ 
full BG  & \checkmark & 24.85   & 35.14  & 1.24 & 3.27  & 5.83  & 9.26  &  3.69  & 18.78  & 0.29  \\  
\midrule
no BG &    & 23.21  & 35.99  & 1.09  & 2.82  & 5.25  & {9.63}  & 3.35  & 20.35 & 0.19 \\
no BG & \checkmark & 25.45   & \underline{38.29}  & 1.33  & 3.79  & 5.93   & \underline{10.63} & 3.98  & \underline{22.23} & 0.29 \\
\midrule
p-label &    & \textbf{25.82}   & \textbf{36.93}  & \textbf{1.51}  & \textbf{3.74} & \textbf{6.08}  & \textbf{10.04}   & \textbf{4.04} & \textbf{21.39} & 0.19 \\
p-label &  \checkmark  & \underline{26.49}  & 37.42 & \underline{1.62}  & \underline{4.09} & \underline{6.27}  & 10.24   & \underline{4.19}   & {21.65} & 0.29\\   
\bottomrule
\end{tabular}
\vspace{-2mm}
\caption{Study on the triplet label assignment. TLA: triplet label assignment, no BG: training without background object label, full BG: assigning background label to all non-foreground entities, p-label: pseudo-label. Notably, in this table, the bolded and underlined values indicate the best results without and with NMS, respectively.}
\label{Tab:labelassign}
\vspace{-4mm}
\end{table*}

\section{Experiments}
\label{sec:experiments}
We conduct experiments on Visual Genome~\cite{vg} and Open Images~\cite{openimage}.
We describe evaluation settings, implementation details, ablation studies and comparisons to the state-of-the-art methods.

\subsection{Datasets and Evaluation Settings}
\label{sec:evaluation_settings}
\textbf{Visual Genome~(VG).} 
VG~\cite{vg} is the most widely used dataset for SGG.
We followed the widely adopted VG split~\cite{Scene_graph_generation_by_iterative_message_passing,tang_treelstm,neural_motifs,Counterfactual_Critic_Multi-Agent_Training_for_SGG} including the most frequent 150 object categories and 50 relation categories.
Since our paradigm generates triplet candidates based on queries, the modes based on ground-truth objects (\eg PredCls and SGCls~\cite{visual_relationship_detection_with_language_priors}) are not suitable here. We adopt the mode of SGDet, which considers both object detection and relation prediction.
The traditional metrics on VG is Recalls~\cite{visual_relationship_detection_with_language_priors}.
Due to the imbalanced class distribution of relations in VG, Recalls are dominated by frequent categories.
Thus, following~\cite{unbias_sgg}, we also utilize mean Recalls~(mR)~\cite{tang_treelstm} and zero-shot Recalls~(zR)~\cite{visual_relationship_detection_with_language_priors} for evaluation.

\begin{table}[t]
\centering
\small
\setlength{\tabcolsep}{2pt}
\begin{tabular}{c|cc|cc|c}
\toprule
Adapt-$\gamma$ & R@20 & R@100  & mR@20  & mR@100 & mR@100 (LA) \\
\midrule
 & 25.58   & 36.57     & {5.93}    & 9.76     &  20.90  \\
\checkmark & \textbf{25.82}   & \textbf{36.93}  & \textbf{6.08}  & \textbf{10.04}  & \textbf{21.39} \\
\bottomrule
\end{tabular}
\vspace{-2mm}
\caption{Study on adaptive focusing parameter.}
\label{Tab:gamma}
\vspace{-3mm}
\end{table}

\textbf{Open Images~(OI).} OI~\cite{openimage} is another large-scale dataset containing annotations for SGG. Currently, two benchmarks for SGG are built on the two versions of this dataset, namely OI~V4 and OI~V6, respectively.
On each benchmark, we carried out experiments and utilized the same backbone as used in~\cite{Bipartite_Graph_imbalance_rel}, and followed their data processing and evaluation metrics.
The training sets and testing sets of OI~V4 contain 54k images and 3k images, respectively. It contains 57 object categories and 9 relation categories.
OI~V6 includes 126k images for training, 2k and 5k images for validation and testing, respectively. It contains 601 object categories and 30 relation categories.
For both OI~V4 and OI~V6, the results are evaluated with the metric of mean Recall@50, Recall@50, weighted mean AP of triplets ($\mathrm{wmAP}_{rel}$), and weighted mean AP of phrase ($\mathrm{wmAP}_{phr}$). The $\mathrm{wmAP}_{rel}$ evaluates the AP of the predicted triplet in which both the subject and object boxes have an IoU of at least 0.5 with ground-truth, while the $\mathrm{wmAP}_{phr}$ uses the union area of the subject and object boxes for IoU calculation.
The final evaluation score is calculated by $\mathrm{score} = 0.2 \times \mathrm{Recall@50} + 0.4 \times \mathrm{wmAP}_{rel} + 0.4 \times \mathrm{wmAP}_{phr}$.

\begin{table*}
\centering
\small
\setlength{\tabcolsep}{3pt}
\begin{tabular}{c|ccc|ccc|ccc|c}
\toprule
Model & R@20 & R@50 & R@100 & zR@20 & zR@50 & zR@100 & mR@20 & mR@50 & mR@100 & speed \\
\midrule
IMP~\cite{Scene_graph_generation_by_iterative_message_passing} & 18.1   & 25.9   & 31.2    & 0.2     & 0.4     & 0.8      & 2.8     & 4.2     & 5.3 & 0.43 \\
G-RCNN~\cite{graphrcnn} & - & 29.7 & 32.8 & - & - & - & - & 5.8 & 6.6 & - \\
VTransE~\cite{vtranse} & 24.5   & 31.3   & 35.5    & -       & 1.9     & 2.6      & 5.1     & 6.8     & 8.0 & 0.40 \\
RelDN~\cite{reldn} & - & 31.4 & 35.9 & - & - & - & - & 6.0 & 7.3 & - \\
GPS-Net~\cite{gps} & - & 31.1 & 35.9 & - & - & - & - & 7.0 & 8.6 & - \\
MOTIFS~\cite{neural_motifs} & 25.1 & 32.1 & 36.9    & -   & 0.1     & 0.2      &  4.1 & 5.5 & 6.8 & 0.45 \\
MOTIFS(Focal)~\cite{neural_motifs} & 24.7 & 31.7 & 36.7 & -   & 0.1     & 0.3      & 3.9 & 5.3 & 6.6 & - \\
MOTIFS(EBM)~\cite{enery_base_loss_sgg} & 24.3 & 31.7 & 36.3 &  0.1 & 0.2 & - & 5.7 & 7.7 & 9.3 & -\\
VCTree~\cite{tang_treelstm} & 24.5   & 31.9   & 36.2    & 0.1     & 0.3     & 0.7      & 5.4     & 7.4     & 8.7 & 0.67 \\
VCTree(EBM)~\cite{enery_base_loss_sgg} & 24.2 & 31.4 & 35.9 &  0.2 & 0.4 & - & 5.7 & 7.7 & 9.1 & -\\
Transformer~\cite{transformer}    & 25.6   & 33.0   & 37.4   & 0.0 & 0.1 & 0.3 & 6.0 & 8.1 & 9.6 & 0.38 \\ 
\hline
VTransE$_\mathrm{TDE}$~\cite{vtranse} & 13.5 & 18.7 & 22.6 & - & 2.0 & 2.7 & 6.3 & 8.6 & 10.5 & - \\
MOTIFS$_\mathrm{TDE}$~\cite{neural_motifs} & 12.4 & 16.9 & 20.3 & - & 2.3 & 2.9 & 5.8 & 8.2 & 9.8 & - \\
VCTree$_\mathrm{TDE}$~\cite{tang_treelstm} & 14.0 & 19.4 & 23.2  & - & 2.6 & 3.2 & 6.9 & 9.3 & 11.1 & - \\
VCTree(EBM)$_\mathrm{TDE}$~\cite{enery_base_loss_sgg} & 14.7 & 20.6 & 24.7  & {1.6} & {2.7} & - & 7.1 & 9.7 & 11.6 & -\\
Transformer$_\mathrm{TDE}^\dagger$~\cite{transformer} & 11.2 & 15.6 & 19.0 & 1.4 & 2.0 & 2.5 & 6.9 & 9.2 & 10.9 & -\\
BGNN~\cite{Bipartite_Graph_imbalance_rel} & - & 31.0 & 35.8 & - & - & - & - & 10.7 & 12.6 & - \\
\midrule
\midrule
Ours  & {25.8} & 32.7  & {36.9} & {1.5} & 2.7 & {3.7} & {6.1} & 8.4 & {10.0} & \textbf{0.19} \\
Ours*  & \textbf{26.1} & \textbf{33.5}  & \textbf{38.4} & {1.5} & 2.7 & {4.0} & {6.2} & 8.6 & {10.3}   & 0.32   \\
\hline
Ours$_\mathrm{TDE}$ & 14.5 & 18.3 & 21.0 & {1.8} & {2.7} & {3.6} & 10.8 & 15.0   & 18.5 & 0.29\\
Ours*$_\mathrm{TDE}$  & 15.0 & 19.7 & 22.9 & {1.6} & {2.7} & {3.8} & {9.8} & {14.6} & {18.0} &  0.54 \\
Ours$_\mathrm{LA}$ & 18.4 & 23.3 & 26.5 & {1.9} & {2.9} & {4.0} & {13.5} & {17.9}   & {21.4} & \textbf{0.19} \\
Ours*$_\mathrm{LA}$  & 18.2 & 23.7 & 27.3 & \textbf{2.0} & \textbf{3.1} & \textbf{4.5} &  \textbf{13.7} & \textbf{18.6} & \textbf{22.5} &  0.32 \\
\bottomrule
\end{tabular}
\vspace{-2mm}
\caption{Comparisons with the state-of-the-art methods at SGDet on Visual Genome (VG). * refers to the 800 queries. LA: logit adjustment~\cite{logit_adjustment}. The reimplemented model is denoted by the superscript $\dagger$. The two blocks indicate the models with or without debiasing techniques, respectively.}
\label{Tab:vg_sota}
\vspace{-2.5mm}
\end{table*}

\subsection{Implementation Details}
\label{sec:implement}
For fair comparison on OI V4, OI V6 and VG, we utilize the same ResNeXt-101-FPN~\cite{fpn,resnext} as the backbone for training Structured Sparse R-CNN.
Our network is optimized by AdamW~\cite{adamw}, and its initial learning rate and batchsize are set to $6.4 \times 10^{-5}$ and 8, respectively. The number of total iterations is $\mathrm{80k}$, and the learning rate is decayed by the factor of 10 on the $\mathrm{47k^{th}}$ and $\mathrm{64k^{th}}$ iterations.
Following~\cite{sparsercnn}, both the triplet detector and the object detector both have 6 detection heads.
Since the classification scores in a triplet share the same weight when inference, the parameters in our loss are set as follows: $\lambda_{{cls}_r}=\lambda_{{cls}_i}=\frac{4}{3}$, $\lambda_{{L1}_i}=5$, $\lambda_{{giou}_i}=2$, $\eta_{{cls}_i}=\frac{1}{3}$, $\eta_{{L1}_i}=\frac{5}{4}$ and $\eta_{{giou}_i}=\frac{1}{2}$.
As for focal loss, we set $\alpha$ and the fixed $\gamma$ to 0.25 and 2, respectively.
We set $\mu$ to 4 for the adaptive focusing parameter.
The number of triplet queries is set to 300 and can be extended into 800.
The number of auxiliary queries is set to 100.
The $\tau$ in logit adjustment is set to 0.3.
Notably, like Sparse R-CNN~\cite{sparsercnn}, NMS can be removed.

\subsection{Ablation Study}
\label{sec:ablation}

We perform ablation studies on VG and report the performance on various Recalls. Furthermore, we report the performance of our models equipped with logit adjustment.

\textbf{Study on the structure modeling.}
We begin our ablation study by exploring the importance of structure modeling in our design.
Specifically, we first propose a purely Sparse R-CNN baseline without explicitly introducing the relation feature vector. This baseline simply treats relation detection as object pair detection without structure modeling and its performance is lower than other variants of Structured Sparse R-CNN, as shown in~\cref{Tab:interaction}.
Then, we investigate the effectiveness of structure modeling module (PF and E2R) in our method by detailed ablations. In~\cref{Tab:interaction}, the results demonstrate that these structured connections are helpful for the relation detection.

\textbf{Study on the triplet label assignment.}
We perform the ablation study on the effectiveness of two-stage triplet label assignment and report the results in~\cref{Tab:labelassign}.
For fair comparison, we report results all with co-training of Siamese Sparse R-CNN, and the only difference is label assignment strategy.
First, we report the results of one-stage triplet label assignment, similar to the training of sparse object detectors, where the object candidates in unmatched triplets are all assigned with the background category (denoted by full BG).
Then, we remove the background label assignment for those objects in unmatched triplets (denoted by no BG).
From the comparison between these two settings, we find that the performance with background supervision at R@20 is better than without it. When the NMS post-processing is used, the performance of the full BG model is poor.
We speculate the background supervision is key to duplicate removal. However, for the objects in triplets with background relations, the full BG model will suppress them indiscriminately though some of them have localized ground-truth objects.
Finally, we compare the previous strategies with our proposed pseudo-label assignment.
Equipped with our pseudo-labels, the overall performance without NMS is better than that of the previous two training strategies, and NMS can still achieve good performance. Among these metrics, its performance on R@20 is consistently the highest. These results demonstrate the effectiveness of our proposed knowledge distillation framework on training our network.
In addition, equipped with NMS, we observe that the absence of background object supervision leads to better performance than utilizing pseudo-labels.
We speculate the noise in pseudo-labels influence the training, thereby resulting in more wrong predictions with low confidence, as well as the lower R@100 and mR@100 performance.

\textbf{Study on the adaptive focusing parameter.} 
We conduct comparative study on the adaptive focusing parameter. In~\cref{Tab:gamma}, the improvement at R@100 and mR@100 shows its effectiveness.

\subsection{Comparisons with the State of the Art}
\label{sec:sota}
\textbf{Visual Genome~(VG).}
We compare our model to the results of the state-of-the-arts methods on VG, shown in~\cref{Tab:vg_sota}.
Following the tradition, we first provide the results of each method on Recalls.
However, VG has a long-tailed distribution of relation categories, and the traditional Recalls are dominated by the frequent categories such as "on".
Accordingly, the mean and zero-shot Recalls are also utilized to evaluate the existing methods~\cite{unbias_sgg}.

The results in~\cref{Tab:vg_sota} show that our Structured Sparse R-CNN achieves the state-of-the-arts performance on multiple metrics.
Specifically, our model shows new state-of-the-arts performance in terms of zero-shot Recalls. As for mean Recalls, our basic model obtains the performance of 8.2\% on average.
Furthermore, our model with 800 queries achieves the best performance on Recalls with an average of 32.7\%.
We speculate that the reason why more queries lead to higher performance lies in the wide range of triplet combinations.
With respect to the processing speed, we conduct experiments on the same server and our model achieves the fastest speed, 0.19 second per image, in the same experimental setting compared to other methods.

Following~\cite{unbias_sgg}, we also report the results of various methods equipped with techniques against long-tailed relation category distribution in~\cref{Tab:vg_sota}.
Our model with TDE~\cite{unbias_sgg} or LA pushes the performance on zero-shot and mean Recalls in the task of SGDet to a new level.
We think it is because the long-tailed relation class distribution limits the performance of models on mean Recalls. With the same debiasing techniques such as TDE, the effectiveness of our design on context feature utilization is revealed.

\begin{table}[t]
\centering
\small
\setlength{\tabcolsep}{1.7pt}
\begin{tabular}{c|ccccc}
\toprule
Model &  mR@50 &  R@50 & $\mathrm{wmAP}_{rel}$ & $\mathrm{wmAP}_{phr}$ & $\mathrm{score}_{wtd}$  \\
\midrule
RelDN~\cite{reldn} & 70.40 & 75.66 & 36.13 & 39.91 & 45.21 \\
GPS-Net~\cite{gps} & 69.50 & 74.65 & 35.02 & 39.40 & 44.70 \\ 
BGNN~\cite{Bipartite_Graph_imbalance_rel} & 72.11 &  75.46 &  37.76 &  41.70 &  46.87 \\
\midrule
Ours  & {72.62} & {74.92} & {43.47} & {48.17} & {51.64} \\
Ours$_\mathrm{LA}$ & \textbf{79.23} & {74.75} & {43.57} & {48.25} & \textbf{51.68} \\
\bottomrule
\end{tabular}
\vspace{-2mm}
\caption{Comparisons with the state-of-the-art methods on Open Images (OI) V4. Following~\cite{Bipartite_Graph_imbalance_rel}, R@50 here is micro-Recall@50~\cite{Gkanatsios_macro_micro_recall}, calculated directly on total ground-truth triplets.}
\label{Tab:oi_sota}
\vspace{-2mm}
\end{table}

\begin{table}[t]
\centering
\small
\setlength{\tabcolsep}{1.7pt}
\begin{tabular}{c|ccccc}
\toprule
Model &  mR@50 &  R@50 & $\mathrm{wmAP}_{rel}$ & $\mathrm{wmAP}_{phr}$ & $\mathrm{score}_{wtd}$  \\
\midrule
MOTIFS~\cite{neural_motifs} & 32.68 & 71.63 & 29.91 & 31.59 & 38.93\\
RelDN~\cite{reldn} & 33.98 & 73.08 & 32.16 & 33.39 & 40.84 \\
VCTree~\cite{tang_treelstm} & 33.91 &  74.08  & 34.16  & 33.11  & 40.21\\
G-RCNN~\cite{graphrcnn} & 34.04 & 74.51 & 33.15 & 34.21 & 41.84\\
GPS-Net~\cite{gps} & 35.26 & 74.81 & 32.85 & 33.98 & 41.69 \\ 
BGNN~\cite{Bipartite_Graph_imbalance_rel} & 40.45 & 74.98 & 33.51 & 34.15 & 42.06\\
\midrule
Ours & {42.84} & {76.66} & {41.47} & {43.64} & \textbf{49.38} \\
Ours$_\mathrm{LA}$ & \textbf{50.73} & {75.70} & {41.14} & {43.24} & {48.89} \\
\bottomrule
\end{tabular}
\vspace{-2mm}
\caption{Comparisons with the state-of-the-art methods on OI V6. Following~\cite{Bipartite_Graph_imbalance_rel}, R@50 here is micro-Recall@50~\cite{Gkanatsios_macro_micro_recall}.}
\label{Tab:oi_sota_v6}
\vspace{-2mm}
\end{table}

\textbf{Open Images~(OI).}
We demonstrate the effectiveness of our method on Open Images and the results are in~\cref{Tab:oi_sota} and~\cref{Tab:oi_sota_v6}.
Consistent with the higher performance at mean Recalls in VG, our method performs better under metrics for each class. When evaluated on R@50, our method still outperforms previous methods.
Moreover, our model with logit adjustment performs slightly worse than the basic one under weighted metrics such as $\mathrm{wmAP}_{rel}$, $\mathrm{wmAP}_{phr}$ and $\mathrm{score}_{wtd}$ in~\cref{Tab:oi_sota_v6}, with the drop on R@50.

\begin{figure}[t]
\centering
\includegraphics[width=1.0\linewidth]{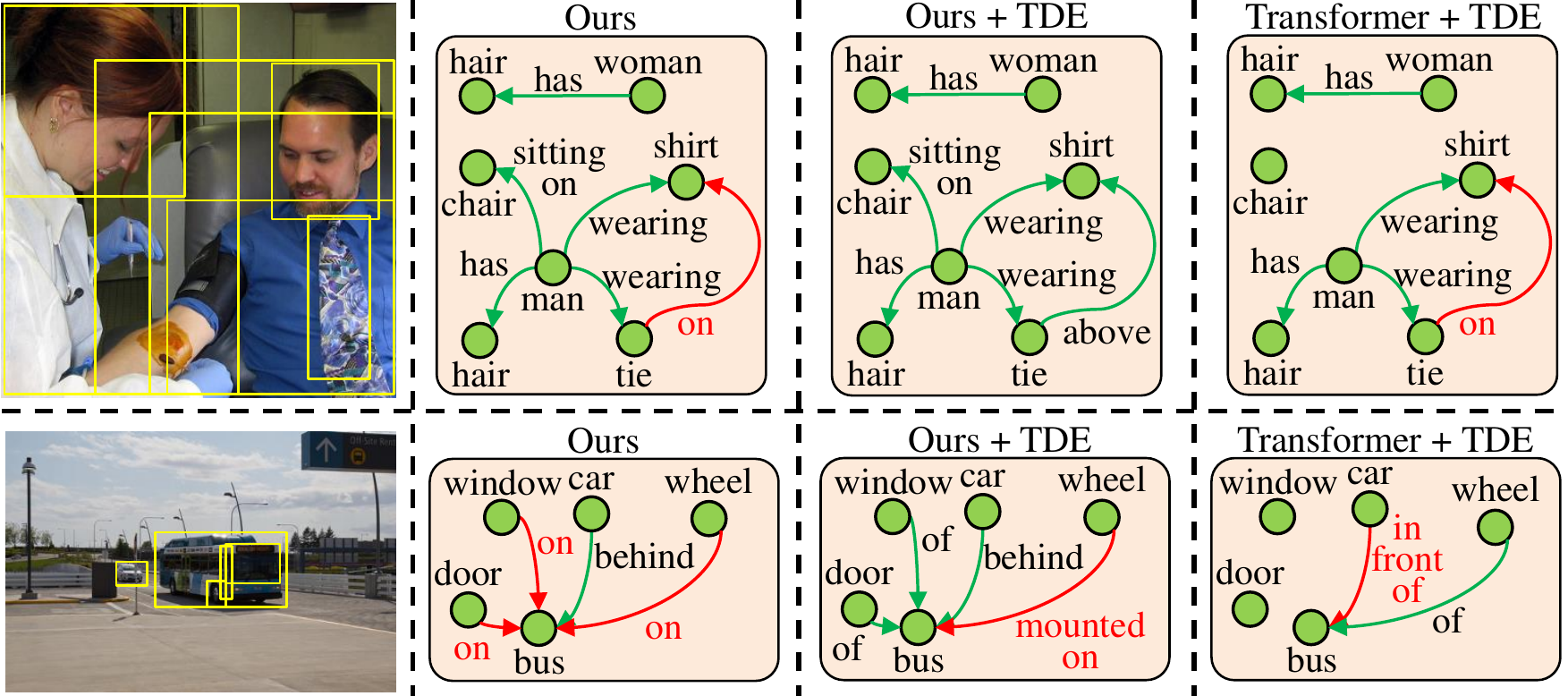}
\vspace{-4mm}
\centering
   \caption{Results of Recall@100 from our model and another method. Due to space limitation, only the directed edges matching the ground-truth pairs are presented. The misclassified relations are marked in red.}
\label{Fig:quality}
\end{figure}

\textbf{Qualitative analysis.}
We visualize the detection results of SGG on VG in~\cref{Fig:quality}.
In general, comparing the results between the last two column, we see that our model detects more correct relation prediction than previous state-of-the-art method, which demonstrates the effectiveness of our method.

\section{Conclusion}
In this paper, we have presented a new perspective on scene graph generation (SGG) as a direct set prediction problem, and proposed a simple, sparse, and unified framework for SGG, termed as Structured Sparse R-CNN. The key to our method is a set of learnable triplet queries and a structured triplet detector, which could be jointly optimized in an end-to-end manner. In addition, we present a relaxed and enhanced training strategy based on the knowledge distillation from a Siamese Sparse R-CNN. We also propose to use adaptive focusing parameter and logit adjustment for imbalance data distribution. We perform experiments on several datasets: Visual Genome and Open Images V4/V6, and the results demonstrate that our method achieves the state-of-the-art performance.

\paragraph{\bf Acknowledgements.} {This work is supported by National Natural Science Foundation of China  (No.62076119, No.61921006),  Program for Innovative Talents and Entrepreneur in Jiangsu Province, and Collaborative Innovation Center of Novel Software Technology and Industrialization.}

\newpage

\appendix

\section*{Appendix}

\section{Technical Details of Modules}

Dynamic convolution~\cite{sparsercnn} is used for the feature extraction. We first perform the dynamic convolution on the subject box and object box with the relation vector to obtain the relation feature. Then, we utilize the same dynamic convolution to perform convolution on the union of the two boxes with the relation feature. The specific process for each dynamic convolution is as follows:
\begin{equation}
\begin{gathered}
\mathcal{F}_1 = \sum_{i=1}^{d_r}{ x_i \cdot W^{(1)}_i}, 
V_1 = \mathrm{ReLU}(\mathrm{LN}(\mathrm{Conv_{1 \times 1}}(V_0, \mathcal{F}_1))),\\
\mathcal{F}_2 = \sum_{i=1}^{d_r}{ x_i \cdot W^{(2)}_i}, 
V_2 = \mathrm{ReLU}(\mathrm{LN}(\mathrm{Conv_{1 \times 1}}(V_1, \mathcal{F}_2))),\\
v_2 = \mathrm{Flatten}(V_2), y = \mathrm{LN}(x + \mathrm{ReLU}(\mathrm{LN}(W_v v_2 ))), 
\end{gathered}
\label{Equ:dynamic_head}
\end{equation}
where $x \in \mathbb{R}^{d_r}$ denotes one relation vector.
$V_0 \in \mathbb{R}^{{C} \times H \times W}$ is the flattened features from roi pooling.
$W^{(1)}_i\in \mathbb{R}^{K_1 \times C}$, $W^{(2)}_i \in \mathbb{R}^{C \times K_1}$ and $W_{v} \in \mathbb{R}^{d_r \times CHW}$ are linear transformation matrices. $K_1$ denotes the number of filters.
$\mathrm{Flatten}(\cdot)$ means reshaping a matrix to the vector form.
$\mathrm{Conv_{h \times w}}(A, B)$ means $h \times w$ convolution on feature map $A$ with filters $B$.
Bias terms are ignored.
Furthermore, since the input $V_0$ for each dynamic convolution has a fixed size (e.g. $H = W = 7$), we can take $W_v$ as a special case of $H \times W$ convolutions. Thus, following~\cite{mobilenetv2}, we decouple $W_v$ into depthwise $H \times W$ convolutions and $1 \times 1$ convolutions with intermediate expansion to high dimensions (e.g. 2048).
Since the dimensionality of object features is larger than the channel of feature maps, this operation can reduce the quantity of parameters.  

As for the relation classification, we adopt the multi-branch structure similar to~\cite{unbias_sgg,reldn} and utilize the fusion operator in~\cite{gps}. The specific process is as follows:
\begin{equation}
\begin{gathered}
G_{s}=\mathrm{LN}({W^s_{r_1}} F_s ), ~~~~ G_{o}=\mathrm{LN}({W^o_{r_1}} F_o ), \\
G_{so} = \mathrm{ReLU}(G_{s} + G_{o}) - (G_{s} - G_{o}) \odot (G_{s} - G_{o}), \\
f'_r = {W^r_{cls}}\mathrm{ReLU}(G_{so}), ~~~~
f''_{r} = f_r + f'_r,
\end{gathered}
\label{Equ:rel_cls}
\end{equation}
where $f_r$ is the logit calculated from the main part of our relation detection.
$f''_{r}$ is the final logit for relation classification.
${W^s_{r_1}}, W^o_{r_1} \in \mathbb{R}^{d \times d}$ and $W^r_{cls}  \in \mathbb{R}^{N_{cls} \times d}$ are linear transformation matrices.
$\odot $ represents the element-wise multiplication.

\section{Two-stage Triplet Label Assignment}

\begin{figure*}
\centering
\includegraphics[width=0.7\linewidth]{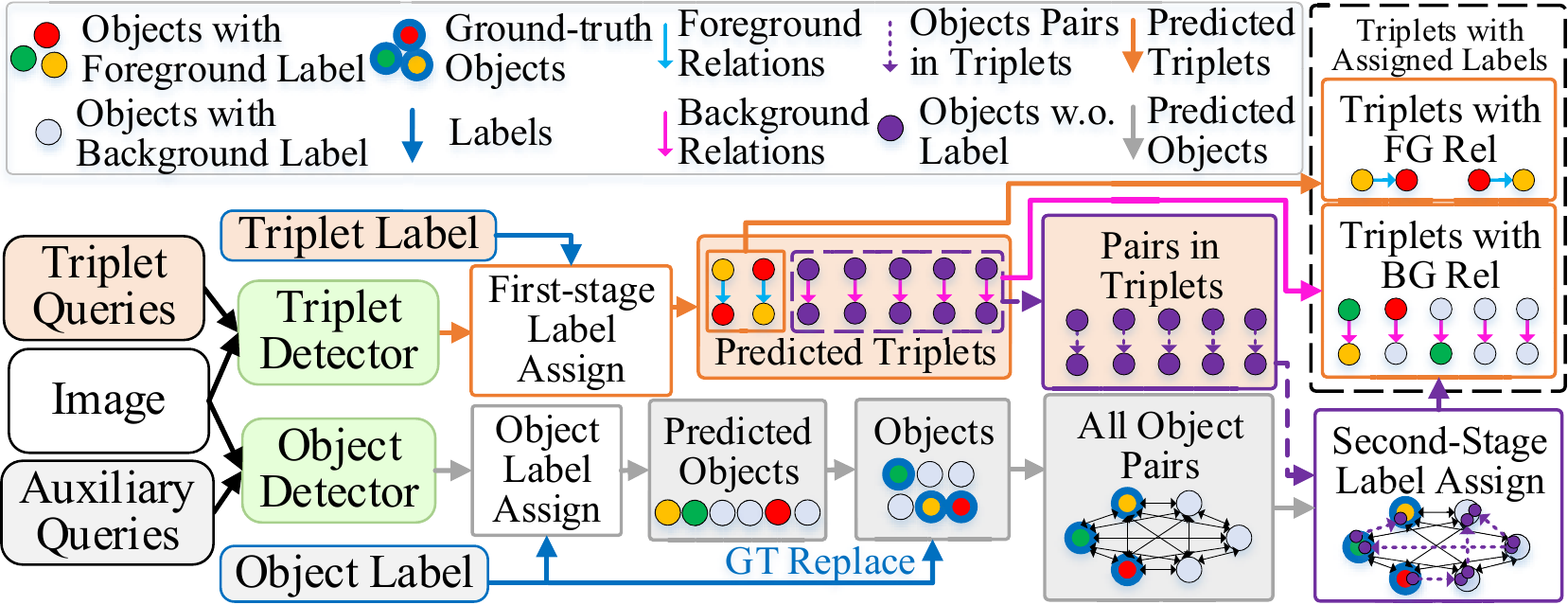}
\caption{The procedure of two-stage triplet label assignment.}
\label{Fig:distill}
\end{figure*}

The triplet detector includes both object pair detection and relation recognition. Its performance heavily depends on object pair detection component. However, the existing SGG benchmarks often contain sparse annotations and fail to cover all object pair instances.
To increase the recall of object pair detection, we need to generate some virtual object pairs as pseudo-labels.
Thus, we use an extra object detector to yield a set of detected boxes, serving as the candidates to form virtual object pairs.
Even though these generated object pairs are not in the SGG ground-truth, they could be used to train the object pair detection component under the object bounding box and classification loss.

As shown in~\cref{Fig:distill}, the procedure of two-stage triplet label assignment is as follows:
{\em first}, the object label assignment is conducted between predicted objects and the ground-truth objects, and the first-stage label assignment is conducted between predicted triplets and the ground-truth triplets;
{\em second}, the predicted objects that match the ground-truth in previous label assignment are replaced by the ground-truth. Also, the classification scores of the predicted objects that do not match the ground-truth in the label assignment are replaced by hard-labels of the background, but {\em their boxes are not the replaced};
{\em third}, these objects are organized into a set of object pairs, \ie the pseudo-label set $U$;
{\em forth}, predicted object pairs are directly taken from the predicted triplets that do not match the ground-truth;
{\em last}, the label assignment is carried out between predicted object pairs and pseudo-label set.
Overall, in our method, when assigning ground-truth labels to predicted triplets, instead of using the padding in original set prediction~\cite{detr}, we employ the label assignment with pseudo-labels.

The key to our label assignment is the pseudo-label set $U$. The detections from Siamese Sparse~R-CNN are paired with each other to form the set $U$, thereby making $U$ many elements.
Actually, due to the huge gap between the magnitude of the triplet queries and pseudo-labels ($O(N)$ versus $O(N^2)$), taking $U$ directly for bipartite matching~\cite{hungarian} to get the optimal solution costs too much computation resources.
Therefore, in practice, we adopt an alternative solution to reduce the computation complexity.

Inspired by~\cite{tanjing_detr_temporal_detection}, we first consider a relaxed strategy which takes the first $K$ small matches for each query. After getting the $M \times K$ pseudo-label candidates, where $M$ indicates the number of queries, we remove the duplicates and get $C$ remaining candidates. If $C > M$, we accept these $C$ pseudo-labels as the candidate set for matching, otherwise we increase $K$. In practice, we use a binary search to determine the minimum of $K$, thereby enabling a lower computation complexity than the original algorithm.

The total loss for our whole framework is as follows:
\begin{equation}
\begin{gathered}
\mathcal{L} = \mathcal{L}_{F} + \mathcal{L}_{B} + \mathcal{L}_{obj},
\end{gathered}
\label{Equ:total_loss}
\end{equation}
where $\mathcal{L}_{F}$ and $\mathcal{L}_{B}$ is to train the triplet detector with triplet queries. $\mathcal{L}_{obj}$ is the loss for training Siamese Sparse~R-CNN.

\begin{table*}
\centering
\setlength{\tabcolsep}{3pt}
\begin{tabular}{c|ccc|ccc|ccc}
\toprule
Model & R@20 & R@50 & R@100 &  zR@20 & zR@50 & zR@100 & mR@20 & mR@50 & mR@100 \\
\midrule
MOTIFS$_\mathrm{LA}^\dagger$ & 16.2 & 21.2 & 24.5 & 1.1 & 1.7 & 2.9 & 10.7 & 13.7  & 16.1  \\
Transformer$_\mathrm{LA}^\dagger$ & 16.9 & 21.8 & 24.9  & 1.1 & 2.0 & 3.1 & 10.5 & 13.7   & 16.2  \\
\bottomrule
\end{tabular}
\vspace{-2mm}
\caption{Other methods with post-hoc approach at SGDet on VG. LA: logit adjustment. The reimplemented model is denoted by the superscript $\dagger$.}
\label{Tab:othermodel_la}
\end{table*}

\begin{figure}[t]
\centering
\includegraphics[width=0.8\linewidth]{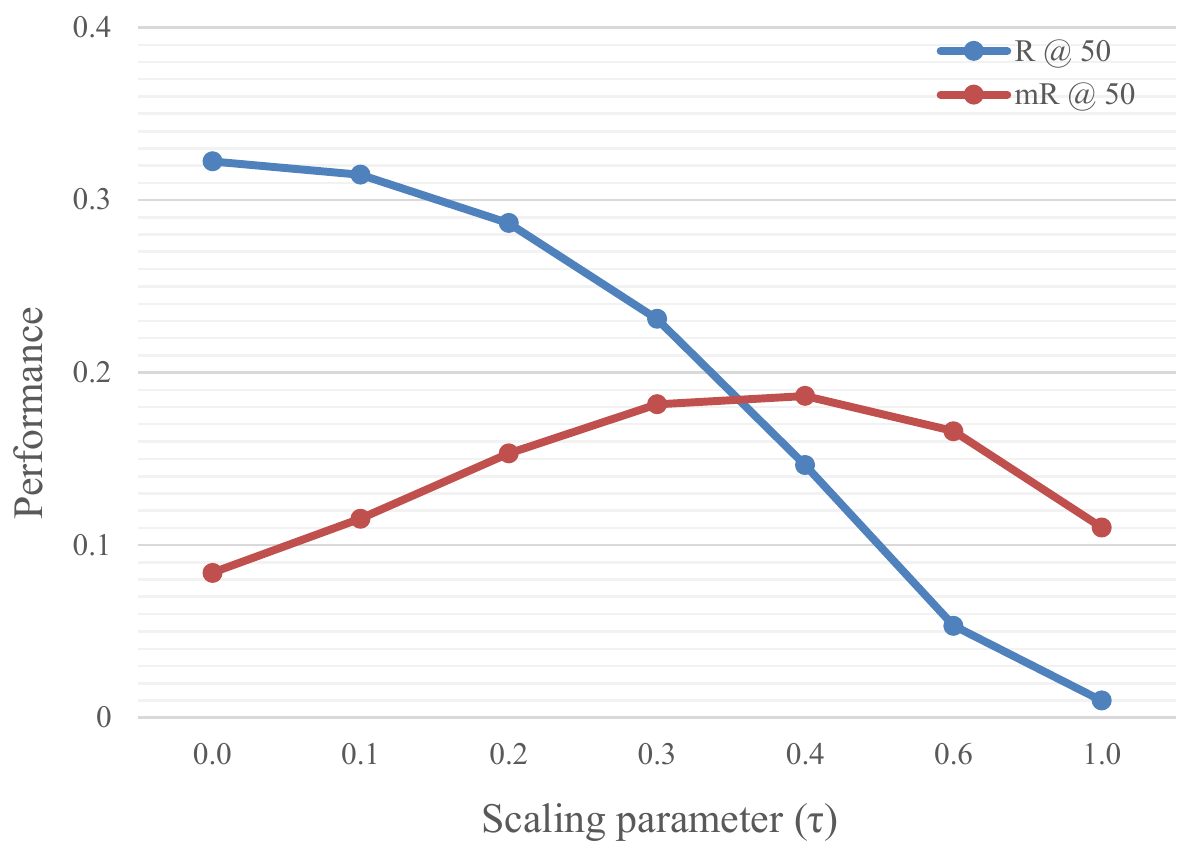}
\vspace{-2mm}
\caption{The performance of our method on R@50 and mR@50 under various scaling parameters.}
\label{Fig:tau_tuning_ours}
\end{figure}

\section{Application of Logit Adjustment}
Logit adjustment~\cite{logit_adjustment} is proposed to tackle long-tailed distribution, and it can be implemented in a post-hoc manner.
The post-hoc logit adjustment is performed by subtracting the logarithm of category frequency multiplied by a tuning parameter $\tau$ from the logit for classification.
Formally, TDE and TE~\cite{unbias_sgg} are similar to this approach.
However, compared with logit adjustment, TDE and TE~\cite{unbias_sgg} lack the temperature scaling $\tau$, which is critical to the performance. Moreover, they calculate statistics for each instance, which costs more resources.

First, we present the selection of $\tau$ for our method. As depicted in~\cref{Fig:tau_tuning_ours}, we find that the best performance on mR@50 is when $\tau$ is between 0.3 and 0.4. However, when $\tau=0.4$, the performance of the head classes is quite poor, and the performance drop on Recall@50 is severe. Therefore, we choose $\tau=0.3$ as the best choice.

Then, we present the performance of other methods equipped with logit adjustment in~\cref{Tab:othermodel_la}. In these methods, we find $\tau=1.75$ is a relatively suitable choice.

\section{Visualization of Model Prediction}
\begin{figure}[t]
\centering
\begin{subfigure}{1.\linewidth}
    \centering
    \includegraphics[scale=0.46]{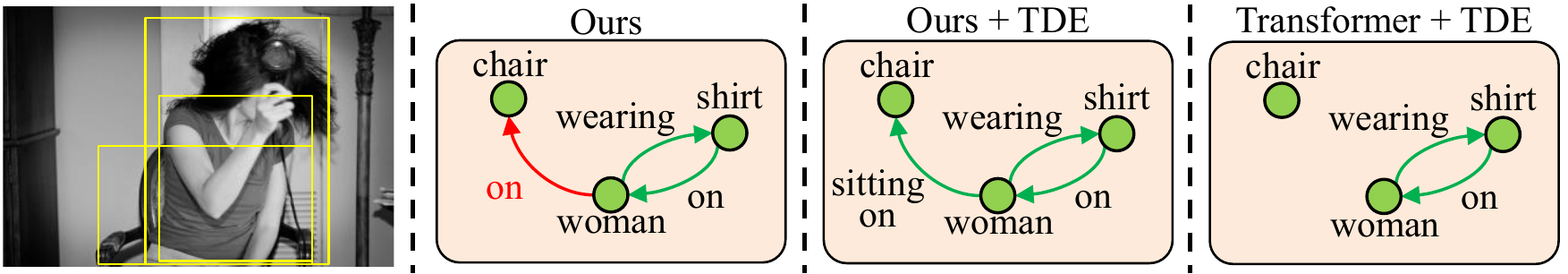}
\end{subfigure}

\centering
\begin{subfigure}{1.\linewidth}
    \centering
    \includegraphics[scale=0.46]{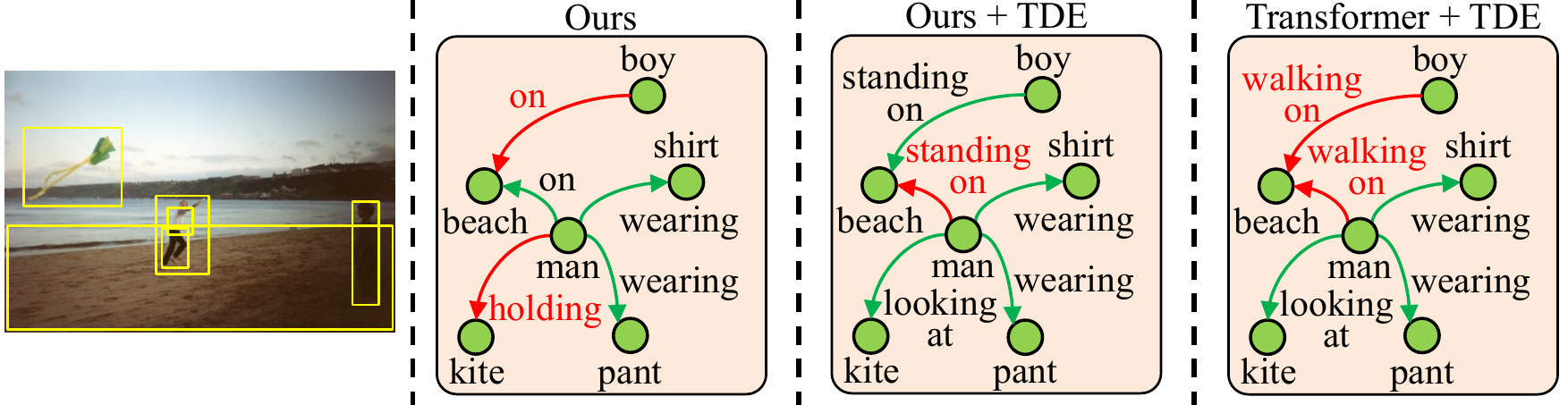}
\end{subfigure}
\centering
   \caption{Results of Recall@100 from our model and another method. We present the directed edges matching the ground-truth pairs, and mark the misclassified relations in red.}
\label{Fig:more_quality}
\end{figure}

Shown in~\cref{Fig:more_quality}, we present more results of our model compared with another method.
We find that our method yields more precise relations where the same entities are detected.
Considering all qualitative analysis in this paper, our framework performs well on various scenes.

\begin{figure}[t]
\centering
\includegraphics[width=1.0\linewidth]{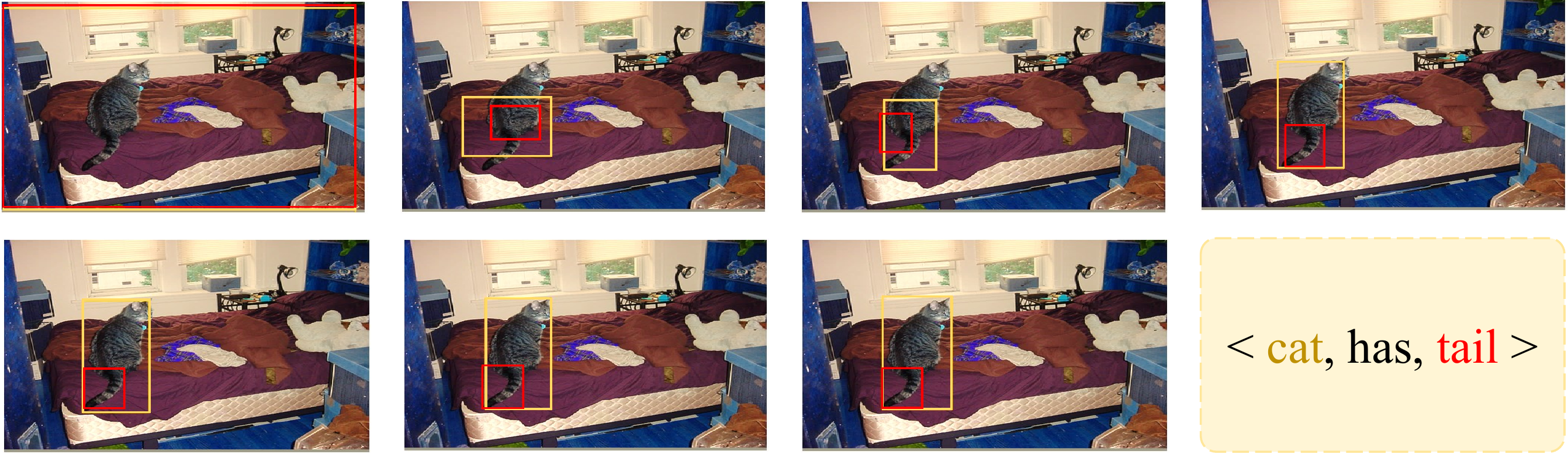}
\vspace{-7mm}
\centering
   \caption{Visualization of an object pair changing with the depth of the triplet detector. The two boxes are marked in orange and red. The direction is from left to right.}
\label{Fig:vis_box}
\vspace{-4mm}
\end{figure}

Moreover, we visualize the change of an object pair in one image. Shown in~\cref{Fig:vis_box}, the initial two learnable boxes almost cover the whole image. As the network gets deeper, the two boxes gradually focus on the main part of the objects (\eg the cat and its tail). Finally, the queries detect the correct object pair with one relation prediction.

\begin{figure}[t]
\centering
\includegraphics[width=0.8\linewidth]{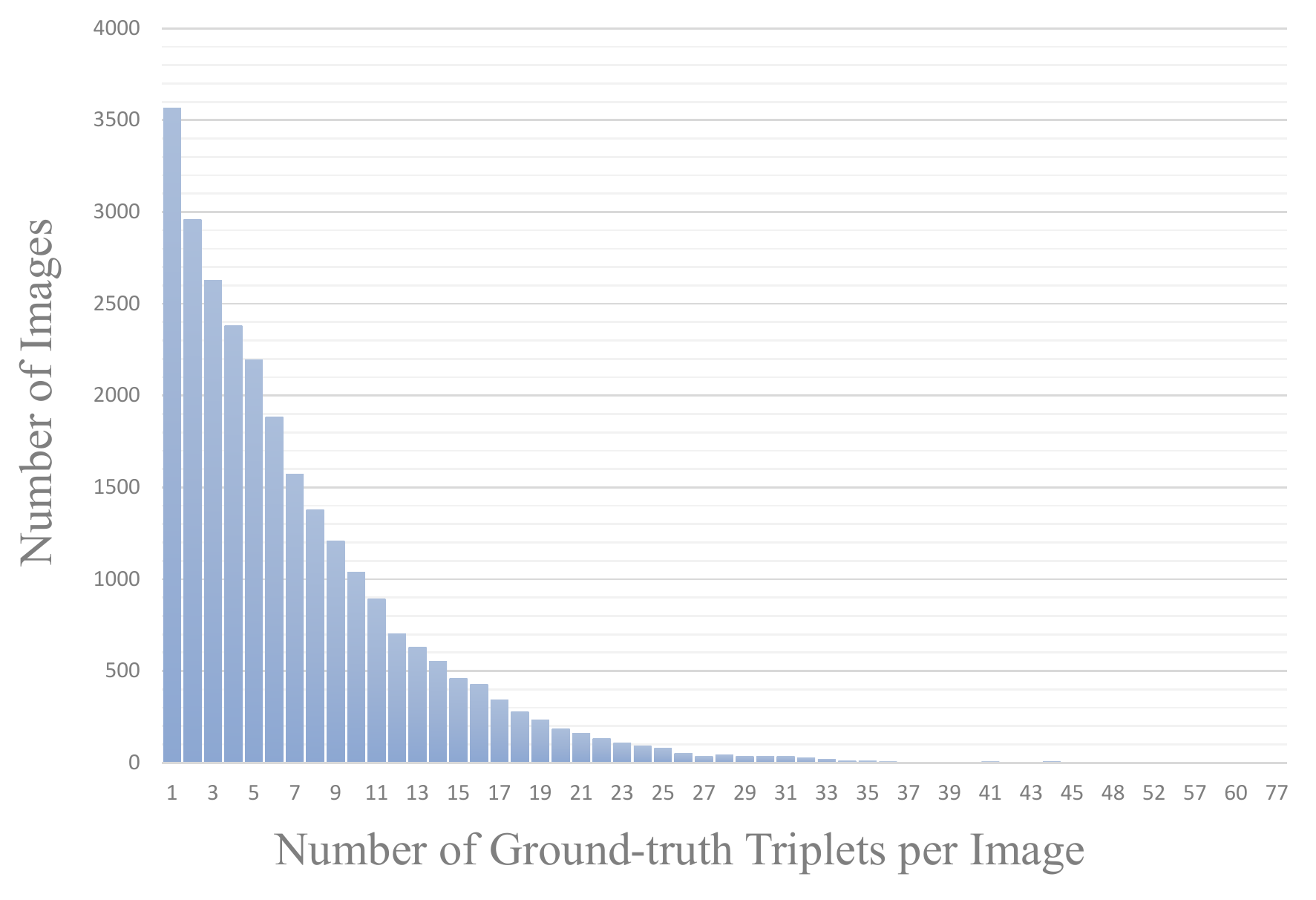}
\vspace{-3mm}
\caption{The relationship between the number of images and the number of ground-truth triplets per image.}
\label{Fig:gt_num_count_perimage}
\end{figure}

\begin{figure*}
\centering
\begin{subfigure}{0.48\linewidth}
    \centering
    \includegraphics[width=0.8\linewidth]{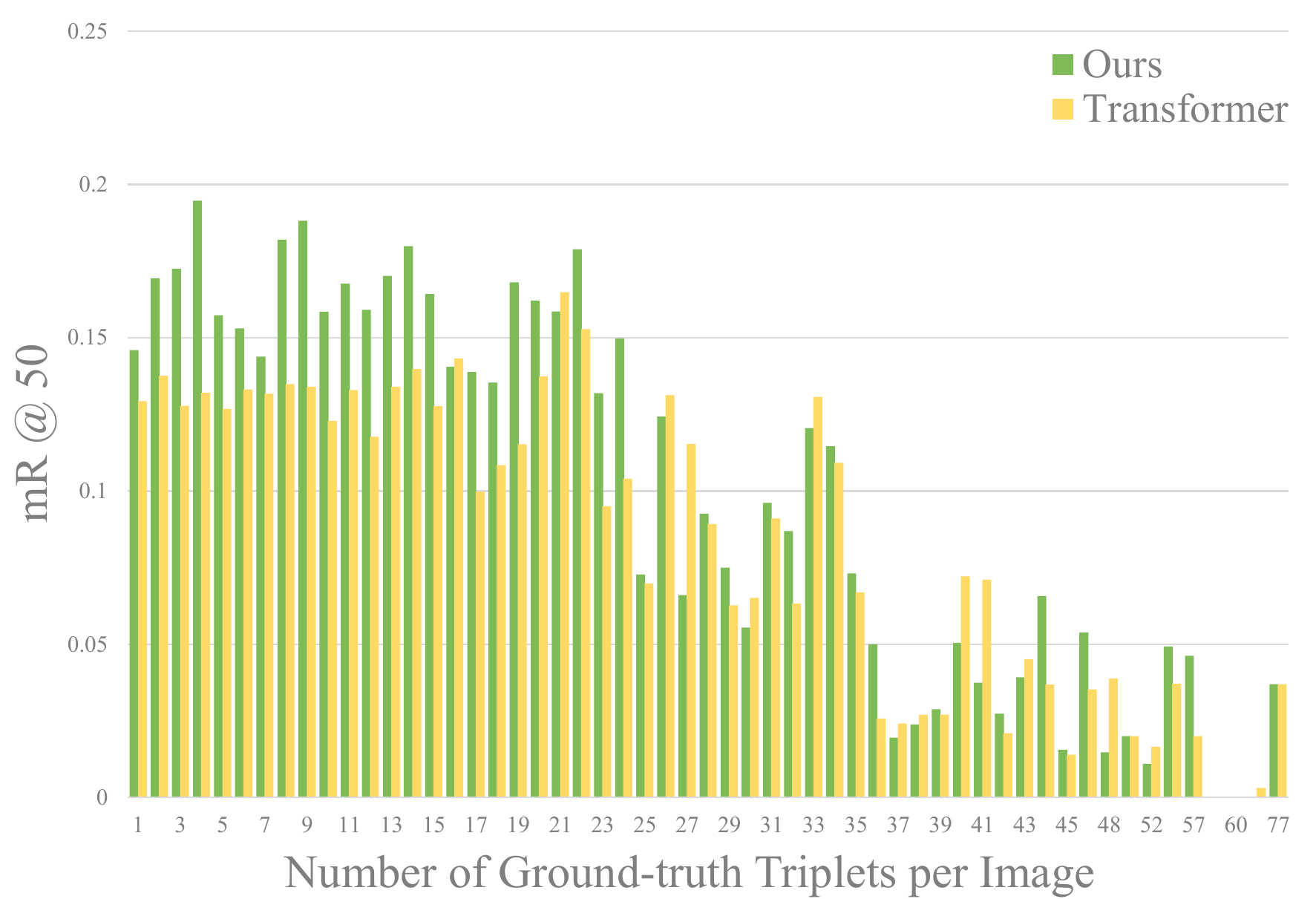}
    \caption{The performance on mean Recall@50.}
    \label{Fig:gtnum_mrecall50}
\end{subfigure}
\centering
\begin{subfigure}{0.48\linewidth}
    \centering
    \includegraphics[width=0.8\linewidth]{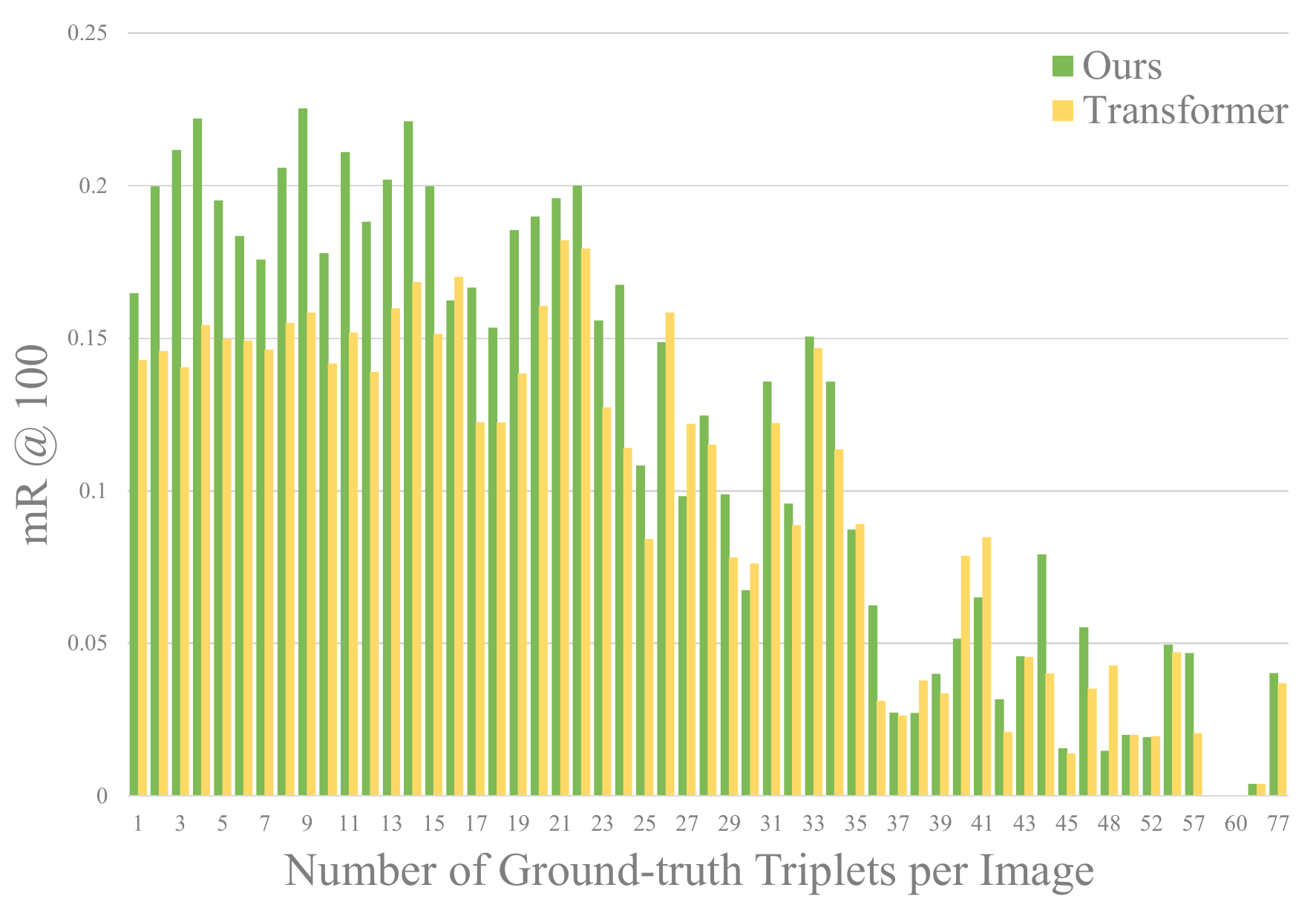}
    \caption{The performance on mean Recall@100.}
    \label{Fig:gtnum_mrecall100}
\end{subfigure}
\centering
\begin{subfigure}{0.48\linewidth}
    \centering
    \includegraphics[width=0.8\linewidth]{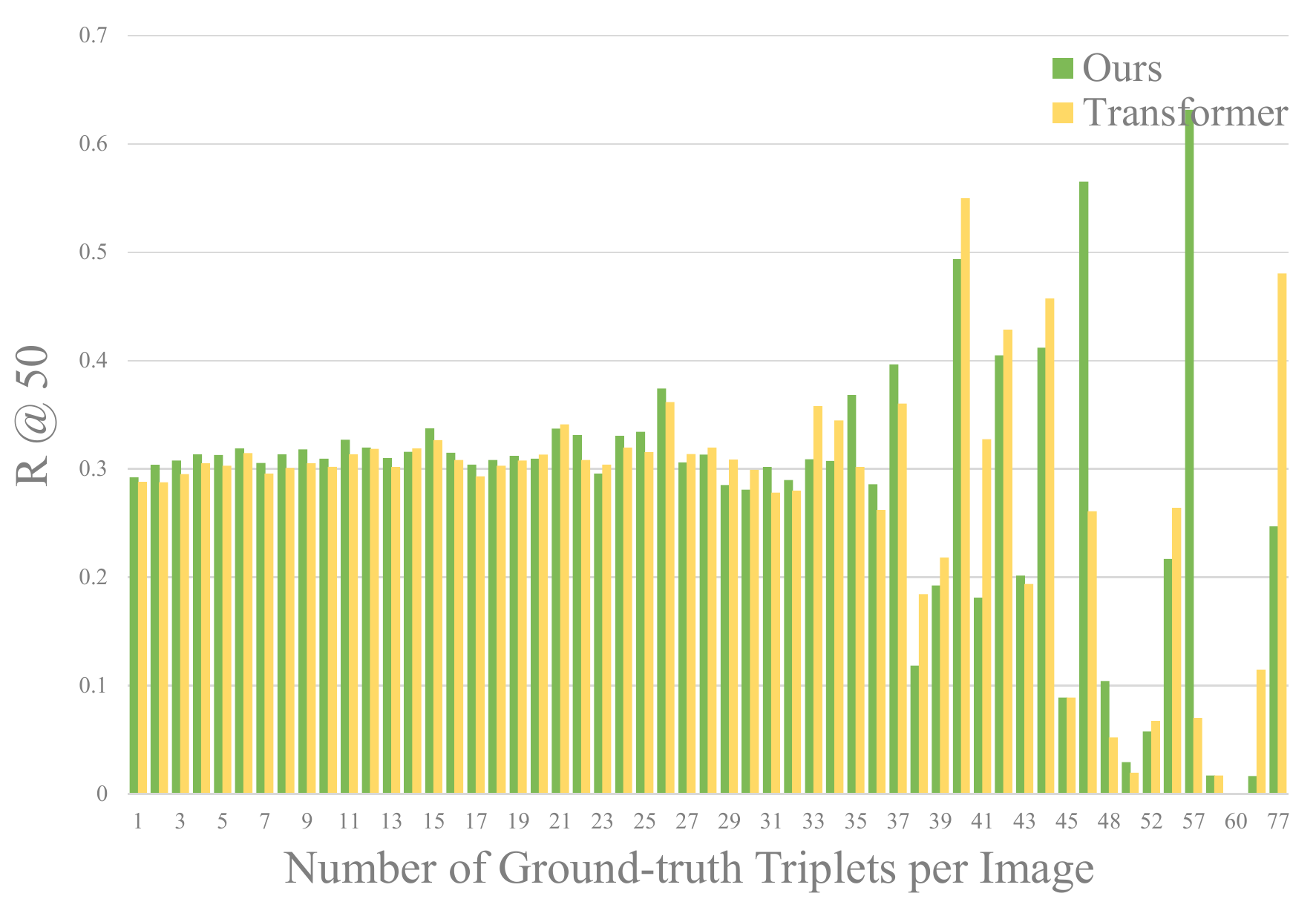}
    \caption{The performance on Recall@50.}
    \label{Fig:gtnum_recall50}
\end{subfigure}
\centering
\begin{subfigure}{0.48\linewidth}
    \centering
    \includegraphics[width=0.8\linewidth]{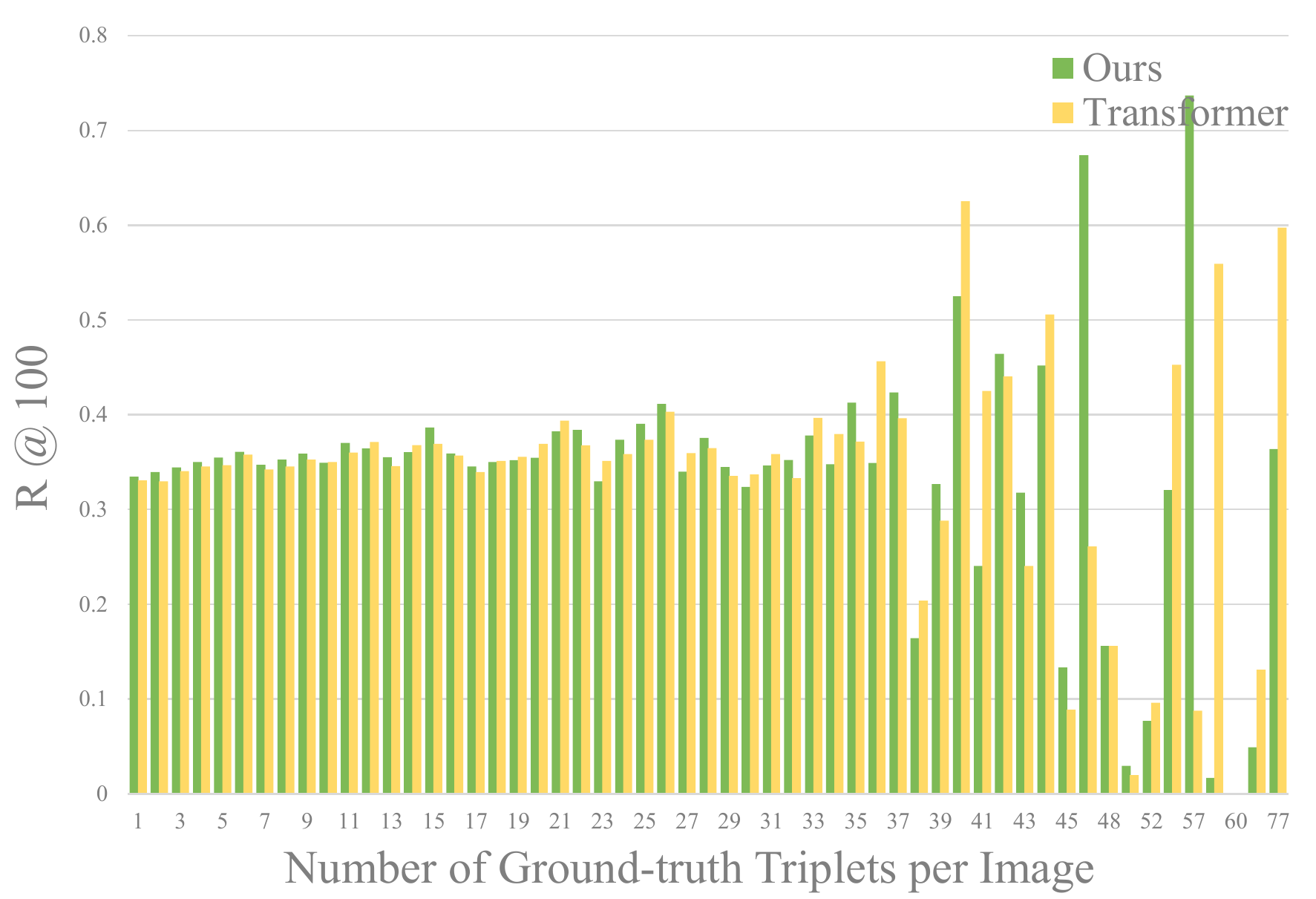}
    \caption{The performance on Recall@100.}
    \label{Fig:gtnum_recall100}
\end{subfigure}
\caption{The performance of our model and transformer.}
\label{Fig:gtnum}
\end{figure*}

\begin{table}[t]
\centering
\begin{tabular}{c|cc}
\toprule
Model  & mR@50 & mR@100 \\
\midrule
Transformer$_\mathrm{LA}$~\cite{transformer}  &  6.0  & 7.0  \\
Ours$_\mathrm{LA}$  &  \textbf{6.3}  & \textbf{7.6}  \\
\bottomrule
\end{tabular}
\vspace{-2mm}
\caption{The average mean Recalls (\%) of various methods with logit adjustment~\cite{logit_adjustment} on images with more than 20 labeled triplets. LA: logit adjustment~\cite{logit_adjustment}.}
\label{Tab:average_mr_20tri}
\end{table}

\begin{table}[t]
\centering
\begin{tabular}{c|cc}
\toprule
Model  & R@50 & R@100 \\
\midrule
Transformer~\cite{transformer}  & 26.2  & {32.9}  \\
Ours  &  {26.9}  & {32.3}  \\
Ours* &  \textbf{28.4}  & \textbf{33.8}  \\
\bottomrule
\end{tabular}
\vspace{-2mm}
\caption{The average Recalls (\%) of various methods on images with more than 20 labeled triplets. * refers to the 800 queries.}
\label{Tab:average_r_20tri}
\end{table}

\begin{figure*}
\centering
\begin{subfigure}{0.48\linewidth}
    \centering
    \includegraphics[width=0.8\linewidth]{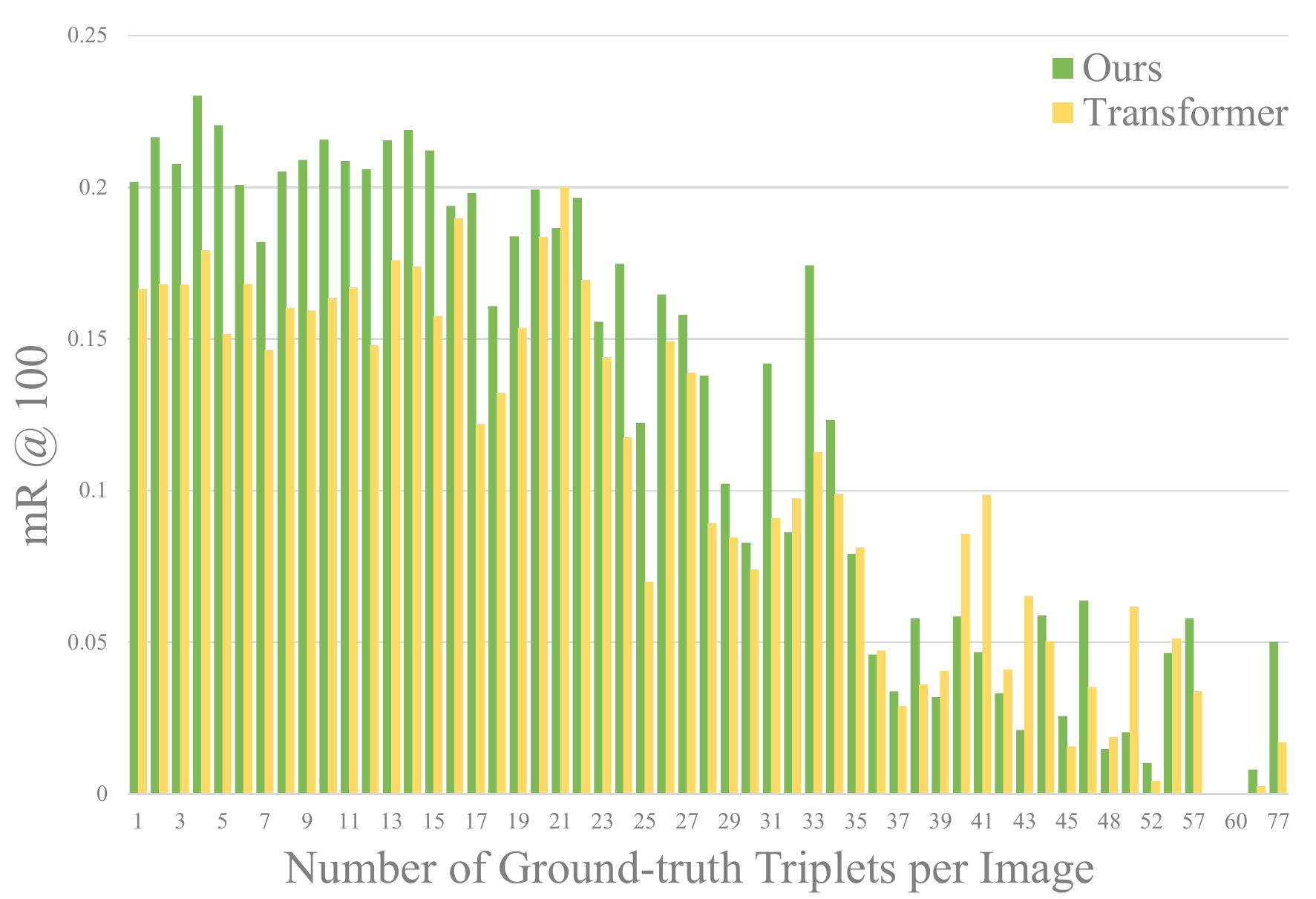}
    \caption{The performance on mean Recall@100.}
    \label{Fig:gtnum_mrecall100_sparsercnn}
\end{subfigure}
\centering
\begin{subfigure}{0.48\linewidth}
    \centering
    \includegraphics[width=0.8\linewidth]{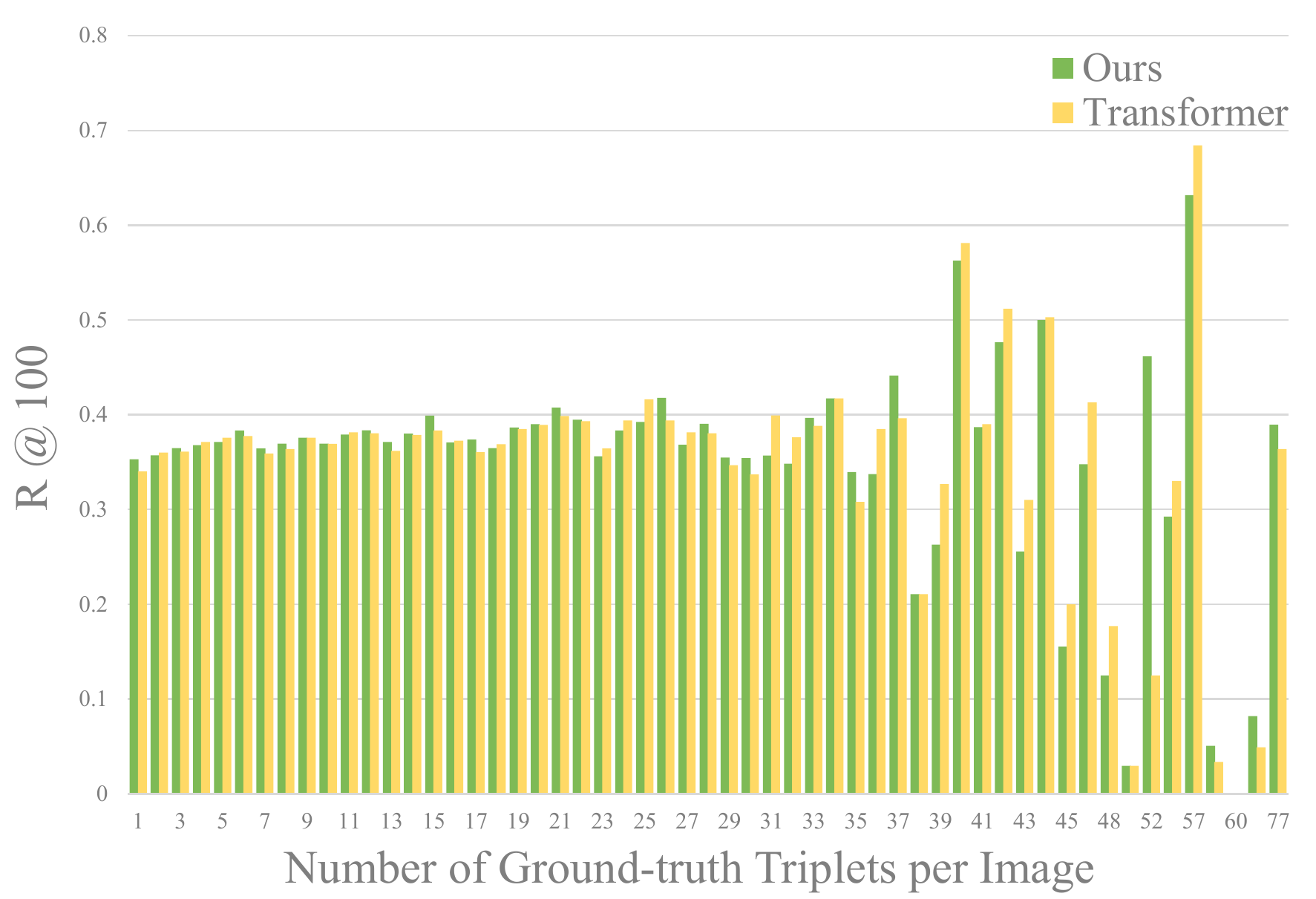}
    \caption{The performance on Recall@100.}
    \label{Fig:gtnum_recall100_sparsercnn}
\end{subfigure}
\caption{The performance of our model and transformer.}
\label{Fig:gtnum_sparsercnn}
\end{figure*}

\begin{table*}[t]
\centering
\setlength{\tabcolsep}{3pt}
\begin{tabular}{cc|cc|cc|c}
\toprule
Model & Object Detector & R@20  & R@100 & mR@20   & mR@100 & Speed \\
\midrule
Transformer$^\dagger$~\cite{transformer} & Sparse R-CNN~\cite{sparsercnn}  & 26.1    & 38.2  & 5.7  & 9.3 & 0.37 \\
Transformer$^\dagger_\mathrm{LA}$~\cite{transformer} & Sparse R-CNN~\cite{sparsercnn}  & 17.1    & 27.2    & 10.5 & 17.3 & 0.37 \\
\midrule
Ours* & - & {26.1}   & {38.4}  & {6.2} & {10.3}   & \textbf{0.32}   \\
Ours*$_\mathrm{LA}$ & - & 18.2  & 27.3  &  \textbf{13.7}  & \textbf{22.5} &  \textbf{0.32} \\
\bottomrule
\end{tabular}
\vspace{-2mm}
\caption{Comparisons with the state-of-the-art methods at SGDet on Visual Genome (VG). * refers to the 800 queries. LA: logit adjustment~\cite{logit_adjustment}. The reimplemented model is denoted by the superscript $\dagger$.}
\label{Tab:vg_sparsercnn_transformer}
\end{table*}

\begin{table}[t]
\centering
\begin{tabular}{cc|c}
\toprule
Object Detector & Context Encoder & $\mathrm{AP_{50}}$ \\
\midrule
Faster R-CNN~\cite{fasterrcnn} & - & 28.1  \\
Sparse R-CNN~\cite{sparsercnn} & - &  29.0  \\
\midrule
Faster R-CNN~\cite{fasterrcnn} &  Transformer~\cite{transformer} & 30.2  \\
Sparse R-CNN~\cite{sparsercnn} &  Transformer~\cite{transformer} &  30.4  \\
\bottomrule
\end{tabular}
\vspace{-2mm}
\caption{The performance (\%) of object detectors on VG~\cite{vg}.}
\label{Tab:obj_det}
\end{table}

\section{More Results and Analysis}
In this part, we will analyze how our method outperforms the state-of-the-art such as transformer~\cite{transformer}.

\subsection{Influence of Annotation Quantity}
\label{Sec:label_quantity}

The mechanism of our method for SGG is based on a limited number of triplet queries. Thus, an intuitive idea is to evaluate the methods on images with different number of ground-truth triplets. As depicted in~\cref{Fig:gt_num_count_perimage}, most images contain fewer than 20 labeled triplets.

Then, we compare our model with transformer with logit adjustment~\cite{logit_adjustment} on mean Recalls~\cite{tang_treelstm} in~\cref{Fig:gtnum_mrecall50} and~\cref{Fig:gtnum_mrecall100}. Consistently, our model outperforms transformer by a great margin on most images with few labeled triplets. As for the images with more than 20 labeled triplets, their performance is not easy to distinguish. Therefore, we calculate the average performance on images with more than 20 labeled triplets. Shown in~\cref{Tab:average_mr_20tri}, our model still outperforms transformer~\cite{transformer} on various mean Recalls with logit adjustment~\cite{logit_adjustment}.

We also compare our model with transformer directly on Recalls~\cite{visual_relationship_detection_with_language_priors} in~\cref{Fig:gtnum_recall50} and~\cref{Fig:gtnum_recall100}. In line with the performance on mean Recalls, our performance on most images with fewer than 20 triplets is slightly better than transformer. Consistently, we calculate the average Recalls of various methods on images with more than 20 labeled triplets in~\cref{Tab:average_r_20tri}. In this table, we find that transformer can outperform our method with a little margin on R@100 when evaluated on images with more triplets. Thus, we evaluate our model with more queries (e.g. 800), and its performance on R@100 is quite better.

\subsection{Influence of Object Detector}
A straightforward idea is investigating how to set the magnitude of queries if both the paradigms achieve similar performance.
Thus, in this part, we compare the transformer~\cite{transformer} with vanilla Sparse R-CNN~\cite{sparsercnn} as the object detector to our method on VG~\cite{vg}.

In~\cref{Tab:vg_sparsercnn_transformer}, we can see that our model with 800 queries can be comparable with the transformer based on vanilla Sparse R-CNN on Recalls. However, our model performs better than transformer on mean Recalls, thereby demonstrating the effectiveness of our relation modeling. Therefore, it is meaningful to design specific structures directly for relation features. Although these structures may cost much computation resources, the sparse modeling on triplets can greatly alleviate this problem. Moreover, the inference speed of our model is faster than that of the dense detector.

Then, we select $\mathrm{AP_{50}}$ for evaluation because all the metrics of SGG utilize an IoU of 0.5 for distinguishing true positive and false positive samples. The performance of object detection is shown in~\cref{Tab:obj_det}. It is obvious that Sparse R-CNN performs better than Faster R-CNN without additional context encoder. However, when both of them are equipped with transformer encoder and used for relation detection, their performance is similar. We speculate that the setting in~\cite{unbias_sgg} forcing the object detector to provide only few high-confident candidates narrows the performance gap between these models. Furthermore, Sparse R-CNN itself depend heavily on context information, and its performance may be similar to that of Faster R-CNN under the same case of utilizing the context.

Consistent with~\cref{Sec:label_quantity}, we also investigate the detailed performance of multi-stage model based on transformer and Sparse R-CNN, and conduct the same comparison with our method with 800 queries. As shown in~\cref{Fig:gtnum_sparsercnn}, the difference between the two method is quite similar with that in~\cref{Fig:gtnum}, which suggests the performance gain of our method is mostly from the images with few annotations.

\section{Societal Impact}
Scene graph generation~(SGG) is a traditional visual scene understanding task and we adopt the open datasets, Visual Genome~\cite{vg} and OpenImage~\cite{openimage}, so there is no negative social impact if the methods in the area of SGG are used properly.

\section{Limitation}
Our method adopts a unified framework for SGG with many fresh operations such as dynamic convolution. Therefore, compared with the previous multi-stage methods, it requires more computation resources for training. 8 RTX 2080ti with 11G GPU memory are necessary for our model with 300 queries. As for training the model with more than 300 queries (\eg 800 queries), 8 Tesla V100 with 32G GPU memory are needed. However, the testing phase only needs 1 RTX 2080ti.

{\small
\bibliographystyle{ieee_fullname}
\bibliography{egbib}
}

\end{document}